\documentclass[10pt, a4paper, onecolumn]{article} 

\usepackage[brazil]{babel} 

\usepackage{float}

\usepackage{microtype} 

\usepackage{amsmath,amsfonts,amsthm} 

\usepackage[svgnames]{xcolor} 

\usepackage[hang, small, labelfont=bf, up, textfont=it]{caption} 

\usepackage{booktabs} 

\usepackage{lastpage} 

\usepackage{graphicx} 

\usepackage[boxruled, linesnumbered, portuguese]{algorithm2e}

\usepackage{blindtext}

\usepackage[colorlinks=true,linkcolor=blue,citecolor=red,urlcolor=black,bookmarks=true,pdfstartview=FitB]{hyperref} 

\usepackage{enumitem} 
\setlist{noitemsep} 

\usepackage{sectsty} 
\allsectionsfont{\usefont{OT1}{phv}{b}{n}} 

\usepackage{geometry} 

\usepackage[T1]{fontenc} 
\usepackage[utf8]{inputenc} 

\usepackage{XCharter} 

\usepackage{fancyhdr} 
\fancypagestyle{firstpage}{ 
	\fancyhf{}
}

\newcommand{\authorstyle}[1]{{\large\usefont{OT1}{phv}{b}{n}\color{DarkRed}#1}} 

\newcommand{\institution}[1]{{\footnotesize\usefont{OT1}{phv}{m}{sl}\color{Black}#1}} 

\usepackage{titling} 

\newcommand{\HorRule}{\color{DarkGoldenrod}\rule{\linewidth}{1pt}} 

\pretitle{
	\vspace{-30pt} 
	\HorRule\vspace{10pt} 
	\fontsize{23}{23}\usefont{OT1}{phv}{b}{n}\selectfont 
	\color{DarkRed} 
}

\posttitle{\par\vskip 15pt} 

\preauthor{} 

\postauthor{ 
	\vspace{10pt} 
	\par\HorRule 
	\vspace{20pt} 
}

\usepackage{lettrine} 
\usepackage{fix-cm}	

\newcommand{\initial}[1]{ 
	\lettrine[lines=3,findent=4pt,nindent=5pt]{
		\color{DarkGoldenrod}
		{#1}
	}{}%
}

\usepackage{xstring} 

\newcommand{\lettrineabstract}[1]{
	\StrLeft{#1}{1}[\firstletter] 
	\initial{\firstletter}\textbf{\StrGobbleLeft{#1}{1}} 
}

\title{ALT\thanks{\url{https://legibilidade.com/}}: um software para análise de legibilidade de textos em Língua Portuguesa} 

\author{
	\authorstyle{Gleice Carvalho de Lima Moreno\textsuperscript{1,3}, Marco P. M. de Souza\textsuperscript{2}, Nelson Hein\textsuperscript{3}, Adriana Kroenke Hein\textsuperscript{3}} 
	\newline\newline 
	\textsuperscript{1}\institution{Departamento de Ciências Contábeis, Universidade Federal de Rondônia, 76801-974, Porto Velho, Rondônia, Brasil}\\
	\textsuperscript{2}\institution{Departamento de Física, Universidade Federal de Rondônia, 76900-726, Ji-Paraná, Rondônia, Brasil}\\
	\textsuperscript{3}\institution{Programa de Pós-Graduação em Ciências Contábeis, Universidade Regional de Blumenau, 89030-903, Blumenau, Santa Catarina, Brasil}\\ 
}

\date{\today} 

\begin{document}

\maketitle 

\thispagestyle{firstpage} 


\lettrineabstract{No estágio inicial da vida humana a comunicação, vista como um processo de interação social, foi sempre o melhor caminho para o consenso entre as partes. O entendimento e a credibilidade nesse processo são fundamentais para que o acordo mútuo seja validado. Mas, como fazê-lo de forma que essa comunicação alcance a grande massa? Esse é o principal desafio quando o que se busca é a difusão da informação e a sua aprovação. Nesse contexto, este estudo apresenta o software ALT, desenvolvido a partir de métricas de legibilidade originais adaptadas para a Língua Portuguesa, disponível na web, para reduzir as dificuldades na comunicação. O desenvolvimento do software foi motivado pela teoria do agir comunicativo de Habermas, que faz uso de um estilo multidisciplinar para medir a credibilidade do discurso nos canais de comunicação utilizados para construir e manter uma relação segura e saudável com o público.}
\newline\newline
\noindent ------------------
\newline

\lettrineabstract{In the initial stage of human life, communication, seen as a process of social interaction, was always the best way to reach consensus between the parties. Understanding and credibility in this process are essential for the mutual agreement to be validated. But, how to do it so that this communication reaches the great mass? This is the main challenge when what is sought is the dissemination of information and its approval. In this context, this study presents the ALT software, developed from original readability metrics adapted to the Portuguese language, available on the web, to reduce communication difficulties. The development of the software was motivated by the  theory of communicative action of Habermas, which uses a multidisciplinary style to measure the credibility of the discourse in the communication channels used to build and maintain a safe and healthy relationship with the public.}


\section{Introdução}


Jürgen Habermas é um filósofo e sociólogo alemão, originário da Escola de Frankfurt, que se empenhou em estudar a democracia se dedicando amplamente à teoria da ação comunicativa publicada em 1981. Essa teoria deu ênfase a forma como a comunicação deve ocorrer, tratando de maneira multidisciplinar a credibilidade na relação entre o sistema (econômico e político) e o mundo da vida (o conhecimento das pessoas). A partir da linguagem inata do indivíduo, ele descreveu que o diálogo deve ocorrer de forma livre, com racionalidade comunicativa e fazendo análise crítica nessa interação para alcançar o essencial e o ápice da comunicação \cite{Habermas}.

Diante disso, influenciados pela teoria da ação comunicativa de Habermas, este trabalho foi desenvolvido para reduzir as falhas que impedem o entendimento das comunicações escritas em Língua Portuguesa. As falhas mais comuns são as faltas de objetividade, de clareza e de simplicidade.

A comunicação é o principal condutor que intervém na busca por uma melhor relação entre pessoas ou grupos de pessoas. Por intermédio do diálogo (forma escrita ou oral), é importante estabelecer uma transmissão de informação respaldada na ética e na moral, a fim de alcançar a persuasão. Seguindo esses critérios, é maior a perspectiva de alcançar a credibilidade e, consequentemente, a concordância da coletividade.

Entretanto, nem sempre é assim. As comunicações escritas são muitas vezes voltadas para um público específico. Dessa forma, o uso de palavras complexas (comuns ao grupo) e de sentenças longas é o mais comum, tornando a leitura difícil e impedindo o entendimento de leitores que fazem parte de outros grupos. Logo, a mensagem não alcança a abrangência esperada, deixando de atender a um maior número de pessoas (seja leigo ou especialista). É quase um jogo de azar com erros e acertos \cite{Gregoire}.

Neste sentido, Habermas constatou, por meio da teoria da ação comunicativa, que os termos usados em uma comunicação devem seguir a quatro pretensões de validade (inteligibilidade, sinceridade, correção normativa e verdade) para alcançar o ápice (credibilidade) nessa etapa tão importante e necessária da interação humana. 

A pretensão da inteligibilidade corresponde ao processo de comunicação realizado com clareza, permitindo assim o fácil entendimento do que foi declarado para se chegar ao consenso entre as partes. De tal modo, essa pretensão se refere ao indicador de compreensibilidade, que ocorre quando o grau de eficácia da comunicação é alcançado. Para medi-lo, há várias métricas de legibilidade textual desenvolvidas ao longo das últimas décadas. Trataremos de algumas delas neste artigo.

Quanto à pretensão da sinceridade, trata-se da divulgação de informações de forma detalhada, apresentando com honestidade o que foi feito ou deixou de ser feito. Para medir de forma precisa esse indicador, devem ser consideradas palavras-chave com o propósito de avaliar se o(s) autor(es) tratou ou trataram de forma pormenorizada o assunto a que se destina o texto.

No que se refere à pretensão da correção normativa, ela tem como princípio o atendimento às regras ou normas estabelecidas, tratando da adequabilidade dos relatórios frente a uma situação específica (ambiental, social, cultural e política, entre outras) ou considerando a capacidade de resposta do leitor ao que se propõe \cite{Baalouch}. Esse indicador em questão é medido por meio das métricas de legibilidade, que apontam o público a que se destina a comunicação.

Já a pretensão da verdade trata da disponibilização de informações confiáveis, considerando sempre a verdade dos fatos. Como dito, a confiabilidade é fundamental para o processo de comunicação, sendo obtida a partir da relação entre diretrizes cumpridas e diretrizes sugeridas com base em padrões preestabelecidos (normas e regras).

A partir da teoria de Habermas, tem-se que, se essas pretensões ou reivindicações estiverem presentes, a comunicação entre o sistema e o mundo da vida será mais sólida.

Isso tem sido visto como um grande desafio para a humanidade, por influenciar o processo de comunicação em suas diversas relações. Como exemplos, podemos citar a relação entre governo e contribuintes; entre empresa e sociedade; entre gestor e colaboradores; entre médico e pacientes; entre cientista e público; e tantos outros casos. Para atenuar as imperfeições no processo de comunicação, em particular na forma escrita, algumas pesquisas receberam destaque por serem pioneiras nesse campo de estudo.

O campo de estudo referido tem relação com a legibilidade, que visa analisar a dificuldade na compreensão de um texto. Alguns estudos, como o de Flesch \cite{Flesh} em 1948, de Gunning \cite{Gunning} em 1952, de Smith e Senter \cite{Smith} em 1967, de Coleman e Liau \cite{Coleman} em 1975, de Kincaid \cite{Flesch-Kincaid} e colaboradores em 1975, e Gulpease \cite{Gulpease} em 1988, além de outros com evidência científica, foram importantes por propor soluções para medir o grau de dificuldade de leitura.

Assim, apresentamos neste trabalho o software ALT: Análise de Legibilidade Textual \cite{ALT}, uma ferramenta desenvolvida para medir índices de legibilidade textual de textos na Língua Portuguesa, usando fórmulas adaptadas para esse idioma a partir das originais. Dentro da teoria da ação comunicativa de Habermas, índices de legibilidade podem ser usados quando o objetivo é obter dados quantitativos das pretensões de inteligibilidade e de correção normativa. Além disso, a pretensão da sinceridade, que busca medir a exaustividade do texto, observando a frequência com que as palavras-chave escolhidas foram mencionadas, também é proposto no software.

Posto isto, o programa ALT foi construído para suprir duas necessidades: 

\begin{enumerate}
	\item Possibilitar a análise de legibilidade textual para textos escritos em Língua Portuguesa.
	
	\item Preencher uma lacuna existente no ambiente científico, uma vez que pesquisadores de diversas áreas desenvolvem estudos com foco na legibilidade textual em Língua Portuguesa e acabam fazendo uso de softwares internacionais baseados em índices de legibilidade não adequados para esse idioma.
	
\end{enumerate}

 Ainda dentro do segundo ponto da lista acima, cabe destacar que mesmo trabalhos recentes, com quatro anos ou menos de publicação, usaram índices de legibilidade textual nos seus idiomas originais (inglês). Dentre esses estudos, podemos citar as referências \cite{ingrid, luciana, guilherme, januario, donizete}, com os quatro primeiros usando o Índice de Legibilidade de Flesch e o último, o Índice Gunning fog. Isso se deve, naturalmente, à ausência de estudos envolvendo a adaptação dos índices de legibilidade estrangeiros para o idioma Português.


Este artigo está organizado da seguinte forma: na Seção \ref{indices-legibilidade} apresentamos uma breve revisão dos índices de legibilidade de interesse, incluindo as suas fórmulas originais. Na Seção \ref{algoritmos} mostramos os algoritmos responsáveis pela contagem de caracteres, de palavras, de sentenças e de sílabas de um texto qualquer. As razões caracteres/palavras, palavras/sentenças e demais combinações são as variáveis-base dos principais índices de legibilidade conhecidos. A adaptação das fórmulas para a Língua Portuguesa é exposta na Seção \ref{formulas-portugues}. A visão geral do programa ALT e suas características são abordadas na Seção \ref{alt}. Com o intuito de conhecer o grau de acurácia do ALT, realizamos na Seção \ref{comparacoes} uma comparação dos índices obtidos pelo ALT com aqueles obtidos pelas fórmulas originais. Por fim, discutimos as limitações da aplicação das fórmulas de legibilidade na Seção \ref{limitacoes} e concluímos este artigo na Seção \ref{conclusoes}.


\section{Os índices de legibilidade}
\label{indices-legibilidade}

Abordamos nesta Seção os índices de legibilidade usados no programa ALT -- Análise de Legibilidade Textual. Com a finalidade de diferenciar daqueles adaptados para a língua portuguesa, que serão apresentados na Seção \ref{formulas-portugues}, faremos referências aos índices abaixo pelos seus nomes originais.

\subsection{ \textit{Flesch reading ease} }
\label{legibilidade-de-flesch}

O \textit{Flesch reading ease} (Índice de Legibilidade de Flesch) é um dos métodos mais antigos capaz de quantificar a ``dificuldade de compreensão'' de textos, conforme as próprias palavras do seu criador, Rudolf Flesch \cite{Flesh}. O método foi desenvolvido em 1943 e a fórmula foi revisada em 1948, tendo sido normalizada em uma escala que varia de 0 (legibilidade mínima) a 100 (máxima legibilidade). A fórmula, voltada apenas para textos na língua inglesa, é dada por

\begin{equation}
	\label{formula-flesch}
	\text{Flesch reading ease} = 206{,}835 - 1{,}015 \times \left( \dfrac{\text{palavras}}{\text{sentenças}}\right) - 84{,}6 \times \left( \dfrac{\text{sílabas}}{\text{palavras}}\right) .
\end{equation}

Como o Índice de Legibilidade de Flesch foi apresentado ao público muitas décadas antes da popularização dos computadores pessoais, o emprego da fórmula era impraticável em artigos longos ou livros. Nesse sentido, o autor recomendava o critério da amostragem: a contagem de sílabas, de palavras e de sentenças podia ser feita a partir de trechos de 100 palavras, sendo de três a cinco textos em artigos e 25 a 30 textos em um livro. A escolha dos textos podia seguir um certo padrão, como começar a partir do terceiro parágrafo de cada página, por exemplo \cite{Flesh}.

Com o advento da popularização dos computadores, a contagem de palavras e de sentenças em textos tão longos quanto os de livros se tornou um processo muito simples. Mas é preciso salientar que a contagem de sílabas ainda continua um desafio. Não há atualmente um algoritmo capaz de fornecer, sem erros, a quantidade de sílabas de uma palavra, e nem mesmo o conceito de sílaba é um consenso entre linguistas \cite{theguardian}.

\subsection{ \textit{Gunning fog index} }
\label{gunning-fog-index}

O \textit{Gunning fog index} (Índice de Nebulosidade de Gunning), muitas vezes traduzido ao pé da letra como Índice de Nevoeiro de Gunning, foi desenvolvido por Robert Gunning em 1952. Gunning, um consultor que chegou a trabalhar para a United Press, o The Wall Street Journal e a Newsweek, foi o primeiro a apresentar uma fórmula de legibilidade que estima os anos de educação formal que uma pessoa deve ter para poder compreender o texto sem dificuldades \cite{Gunning}. Por exemplo, um texto com índice de 12 pontos estaria adequado para um leitor com formação acadêmica de final de Ensino Médio, que gira em torno de 12 anos de estudos (3 anos do Ensino Médio + 9 anos do Ensino Fundamental). Essa escala, baseada em anos de estudos ao invés da escala arbitrária centígrada, é hoje conhecida como nível de instrução ou nível de escolaridade (\textit{grade level}).

A fórmula para a obtenção do nível de instrução do texto é dada por

\begin{equation}
	\text{Gunning fog index} = 0{,}4 \times \left( \dfrac{\text{palavras}}{\text{sentenças}}\right) + 40 \times \left( \dfrac{\text{palavras complexas}}{\text{palavras}}\right) .
\end{equation}

Assim como acontece no Índice de Legibilidade de Flesch, o Índice de Nebulosidade de Gunning é baseado nos conceitos de ``sentenças complexas'' e de ``palavras complexas''. Sentenças complexas, no sentido de Gunning, são aquelas muito longas em termos de quantidade de palavras: observe que a primeira variável (palavras/sentenças) é um dos determinantes no nível de instrução do texto. É intuitivo que grandes sentenças, compostas por muita recursividade \cite{Recursividade}, torna a leitura difícil. Não é raro que mesmo um leitor ou leitora culta tenha que reler uma sentença muito longa para poder compreender a informação contida nela.

Sobre as palavras complexas, Gunning as define como sendo aquelas que contêm três ou mais sílabas. Nomes próprios, jargões familiares e palavras compostas não devem ser levados em conta. É interessante notar a diferença entre o índices de Flesch e de Gunning: enquanto o primeiro engloba um espectro amplo de palavras que vão das muito simples (monossilábicas), passando pelas moderadas (dissilábicas e trissilábicas) e chegando até as muito difíceis (com 4 ou mais sílabas), o segundo destaca que as palavras ou são comuns, ou são complexas. Nesse sentido, o Índice de Nebulosidade de Gunning considera que as palavras ``recado'' e ``heterozigoto'' têm o mesmo grau de complexidade.

\subsection{ \textit{Automated readability index} }

O ARI -- \textit{Automated readability index} -- (Índice de Legibilidade Automatizado) foi desenvolvido por E. A. Smith e R. J. Senter em 1967. Conforme apontado no \textit{paper} seminal \cite{Smith}, seu objetivo foi oferecer um índice de legibilidade para livros, relatórios e manuais técnicos da Força Aérea dos Estados Unidos com o propósito de diminuir o tempo de extração de informação desses documentos.

Até 1967, havia pelo menos três métodos de obtenção de índices de legibilidade textual. Dois deles eram o Índice de Legibilidade de Flesch e o Índice Gunning fog que, conforme já apontado, fazia uso do número de sílabas para inferir a complexidade de uma palavra. O outro era o algoritmo de Dale-Chall \cite{DaleChall}, de 1948, que apresentava o índice em uma escala de 4,9 (ou menos) até 9,9 pontos. O problema deste último era que a fórmula dependia da comparação do texto com uma lista de 763 ``palavras muito comuns'', como \textit{yes} (sim) e \textit{no} (não). Esse índice era bastante interessante para textos infantis, já que o repertório de palavras conhecidas por crianças é pequeno. Entretanto, a fórmula não era adequada para textos voltados ao público adulto. Foi nessa conjuntura que os autores do método ARI apresentaram uma fórmula que evitava a complexidade envolvida na contagem de sílabas do Índice de Legibilidade de Flesch e a presença de uma lista de palavras do Índice de Dale-Chall. O Índice de Legibilidade Automatizado, que é fundamentado na escala de nível de instrução, é dado por

\begin{equation}
	\label{formula-ari}
	\text{ARI} = - 21{,}43 + 0{,}50 \times \left( \dfrac{\text{palavras}}{\text{sentenças}}\right)  + 4{,}71 \times \left( \dfrac{\text{caracteres}}{\text{palavras}}\right).
\end{equation}

\noindent Como é possível perceber, o Índice de Legibilidade Automatizado também é baseado no conceito de ``palavras complexas'' e de ``sentenças complexas''. A vantagem do ARI é a facilidade com que esses conceitos podem ser quantificados, já que basta contar o número de caracteres (\textit{strokes}), de palavras e de sentenças. A contagem dessas três variáveis é bastante simples nos dias atuais com o auxílio de um processador de texto como o Microsoft Word e similares.

\subsection{ \textit{Flesch–Kincaid grade level} }

Em 1975, J. Peter Kincaid e colaboradores recalcularam três índices de legibilidade (ARI, \textit{Gunning fog index} e \textit{Flesch reading ease}) para textos ligados à Marinha dos Estados Unidos. Além disso, foi apresentado também a fórmula de Flesch reescrita na escala de nível de instrução, hoje conhecida como \textit{Flesch-Kincaid grade level} (Nível de Instrução de Flesch-Kincaid):

\begin{equation}
	\label{formula-flesch-kincaid}
	\text{Flesch–Kincaid grade level} = - 15{,}59 + 0{,}39 \times \left( \dfrac{\text{palavras}}{\text{sentenças}}\right) + 11{,}8 \times \left( \dfrac{\text{sílabas}}{\text{palavras}}\right).
\end{equation}

A fórmula de conversão entre as escalas centígrada (de zero a cem) e de nível de instrução, obtida a partir da manipulação das Eqs. (\ref{formula-flesch}) e (\ref{formula-flesch-kincaid}), é dada por

\begin{equation}
	\label{formula-conversao-escalas}
	\text{Flesch–Kincaid grade level} = 63{,}88 - 0{,}38424\times (\text{Flesch reading ease}) - 20{,}7 \times \left( \dfrac{\text{sílabas}}{\text{palavras}}\right).
\end{equation}

\subsection{ \textit{Coleman–Liau index} }

O \textit{Coleman–Liau index} (Índice de Coleman-Liau) segue o critério adotado pelo método ARI, no sentido em que ele foi desenvolvido com o propósito de ser um índice de fácil implementação computacional. Sua fórmula, criada por Meri Coleman e T. L. Liau, é dada por \cite{Coleman}

\begin{equation}
	\label{formula-coleman-liau}
	\text{Coleman–Liau grade level} = -15{,}8 - 2{,}96 \times \left( \dfrac{\text{sentenças}}{\text{palavras}}\right) + 5{,}88 \times \left( \dfrac{\text{letras}}{\text{palavras}}\right).
\end{equation}

\noindent Esse índice é bastante semelhante ao ARI. A diferença mais notável é inferir a complexidade das sentenças pela razão sentenças/palavras, o que é o inverso do que aparece no ARI e nos outros índices já apresentados (palavras/sentenças).

\subsection{ \textit{Indice Gulpease} }

Desenvolvido pelo \textit{Gruppo Universitario Linguistico Pedagogico} (GULP) da Universidade de Roma La Sapienza em 1987, o \textit{Indice Gulpease} (Índice Gulpease) fornece um número para a legibilidade de textos na língua italiana. Também delimitado pela escala centígrada, sua fórmula é dada por

\begin{equation}
	\label{formula-gulpease}
	\text{Indice Gulpease} = 89 + 300 \times \left( \dfrac{\text{sentenças}}{\text{palavras}}\right) - 10 \times \left( \dfrac{\text{letras}}{\text{palavras}}\right).
\end{equation}

Esse índice também não usa o critério do número de sílabas para delimitar as palavras complexas.


\section{Os algoritmos}
\label{algoritmos}

\begin{figure}[H]
	\centering
	\includegraphics[width=0.4\linewidth]{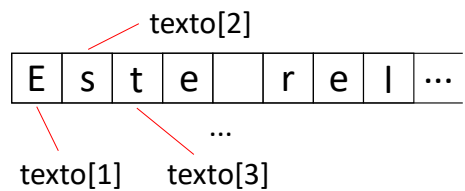}
	\caption{Armazenamento do conteúdo de um documento que começa com ``Este relatório apresenta ...'' no vetor \texttt{texto}. Nesse exemplo, as primeiras cinco componentes do vetor são os caracteres \texttt{E}, \texttt{s}, \texttt{t}, \texttt{e} e um espaço em branco. Uma $k$-ésima componente arbitrária do vetor é representada por \texttt{texto[k]}.}
	\label{fig1}
\end{figure}

Para contar a quantidade de caracteres, de palavras, de sentenças e de sílabas no software ALT, o primeiro procedimento é armazenar todos os caracteres do documento em um vetor, que chamaremos de \texttt{texto}, de $N$ componentes, conforme exemplo ilustrado na Fig. \ref{fig1}. Esses caracteres podem incluir letras, algarismos, sinais de pontuação, outros símbolos dispostos no teclado e demais símbolos encontrados em outros idiomas. A contagem dessas quatro variáveis mencionadas é feita a partir da análise das componentes do vetor \texttt{texto}, cujos algoritmos são descritos nas seções abaixo.

\subsection{Contagem de caracteres}
\label{contagem-caracteres}

Consideramos como caractere todas as letras, tanto maiúsculas como minúsculas, e o símbolo \texttt{-} (hífen), além de números, sinais e outros símbolos. Definimos então a variável \texttt{qntCaracteres}, que representa o total de caracteres do texto. Conforme as componentes do vetor \texttt{texto} são lidas, uma função é chamada para saber se o símbolo armazenado no vetor é uma letra ou hífen. Em qualquer um dos casos, a variável \texttt{qntCaracteres} é incrementada em uma unidade.

Ao final da leitura das $N$ componentes do vetor \texttt{texto}, a variável \texttt{qntCaracteres} conterá o total de caracteres do documento.


\subsection{Contagem de palavras}

Para a contagem de palavras, a ideia inicial está contida em \cite{Smith}: incrementar em uma unidade a variável \texttt{qntPalavras} cada vez que um espaço vazio (produzido pela barra de espaço do teclado) é encontrado no vetor \texttt{texto}. Essa é uma solução simplificada, mas não inteiramente correta. Como os espaços estão entre as palavras, há, em geral, um espaço vazio a menos do que o total de palavras de uma sentença. E devemos considerar também o caso dos parágrafos. Três parágrafos com apenas uma palavra cada não contêm nenhum espaço vazio.

O algoritmo integrado no ALT é então dado pela seguinte instrução: incremente a variável \texttt{qntPalavras} sempre que a componente \texttt{texto[k]} for a última do vetor, ou um espaço vazio (\texttt{`` ''}), ou um retorno de carro, produzido pela tecla \texttt{Enter}/\texttt{Return} (\texttt{``$\backslash$n''}), ao mesmo tempo em que a componente anterior (\texttt{texto[k-1]}) não seja nenhum dos objetos anteriores e nem o símbolo hífen.

\subsection{Contagem de sentenças}

Para contarmos as sentenças, incrementamos a variável \texttt{qntSentencas} sempre que a componente \texttt{texto[k]} for igual a um ponto (\texttt{.}), ou um ponto de exclamação (\texttt{!}), ou um ponto de interrogação (\texttt{?}), ou um ponto e vírgula (\texttt{;}) ao mesmo tempo em que \texttt{texto[k-1]} não seja nenhum desses sinais de pontuação, para que não sejam contabilizadas sentenças a mais em casos onde esses sinais de pontuação aparecem em sequência. O ponto e vírgula (\texttt{;}) foi considerado aqui pela frequência em que são exibidos para citar listas de itens, de ideias, de separação de orações coordenadas e outros casos, o que representaria uma sentença longa, interferindo no índice final obtido.

\subsection{Contagem de sílabas}

O ponto de partida é um peculiaridade da língua portuguesa: o total de sílabas de uma palavra é igual à quantidade de vogais contidas nela. O problema consiste então em eliminar da contagem todas as semivogais \texttt{i} e \texttt{u} que podem aparecer nos ditongos e tritongos. Esta não é uma tarefa simples e o que apresentamos aqui é apenas um algoritmo que devolve o número \textit{aproximado} de sílabas de uma palavra.

A primeira etapa consiste em armazenar todas as vogais, ditongos e tritongos conhecidos, incluindo a presença de eventuais acentos, nos vetores \texttt{vogal}, \texttt{ditongo} e \texttt{tritongo}:

\begin{algorithm}[ht]
	vogal = [a, ã, â, á, à, e, é, ê, i, í, o, ô, õ, ó, u, ú]
	
	ditongo = [ ãe, ai, ão, au, ei, eu, éu, ia, ie, io, iu, õe, oi, ói, ou, ua, ue, uê, ui ]
	
	tritongo = [uai, uei, uão, uõe, uiu, uou]
\end{algorithm}

Em seguida, incrementamos a variável \texttt{qntSilabas} toda vez em que \texttt{texto[k]} é igual a qualquer uma das componentes do vetor \texttt{vogal} ou da sua correspondente em maiúscula. O último passo é desconsiderar as semivogais dos ditongos e tritongos. 

Para o caso dos ditongos, começamos reduzindo a variável \texttt{qntSilabas} em uma unidade toda vez em que \texttt{texto[k-1]} + \texttt{texto[k]} é igual a qualquer uma das componentes do vetor \texttt{ditongo} ao mesmo tempo em que \texttt{texto[k-2]} é uma consoante. 

Para eliminar as semivogais dos tritongos, reduzimos a variável \texttt{qntSilabas} em uma unidade toda vez em que \texttt{texto[k-2]} + \texttt{texto[k-1]} + \texttt{texto[k]} for igual a qualquer uma das componentes do vetor \texttt{tritongo}.

O procedimento descrito neste algoritmo é capaz de retornar corretamente o número de sílabas de aproximadamente 96\% das palavras.

\section{Adaptação das fórmulas para a Língua Portuguesa}
\label{formulas-portugues}

Antes de tudo, é interessante observar as seis fórmulas de legibilidade usadas no software ALT, equações (1)-(6). Todas elas dependem linearmente de duas variáveis, de forma que elas podem ser representadas genericamente por

\begin{equation}
	\label{plano}
	f(x,y) = C_1 + C_2 \;x + C_3 \;y.
\end{equation}

\noindent A variáveis estão sendo representadas por $x$ e $y$, que podem ser palavras/sentenças ou sílabas/palavras, por exemplo, e $C_k$ é o $k$-ésimo coeficiente. Isso significa que, a partir de uma regressão linear múltipla, é possível obter o plano dado pela equação (\ref{plano}) que melhor se ajusta a um certo conjunto de pontos. Nesse momento vale enfatizar o porquê de não utilizarmos outras fórmulas de legibilidade no software ALT, como o nível SMOG de G. Harry McLaughlin (\textit{SMOG grade}) \cite{Smog},

\begin{equation}
	\label{smog}
	\text{SMOG grade} = 1{,}0430 \sqrt{ 30\times \left( \dfrac{\text{polissílabos}}{ \text{sentenças} }\right)  } + 3{,}1291,
\end{equation}

\noindent onde polissílabos são o número de palavras com três ou mais sílabas. Uma dependência não-linear entre o nível de instrução e a variável polissílabos/sentenças torna o ajuste da superfície dada pela Eq. (\ref{smog}) mais difícil: enquanto que apenas três pontos determinam um plano, são necessários infinitos pontos para caracterizar qualquer outra superfície.

Adaptar para a Língua Portuguesa as fórmulas dadas pelas equações (\ref{formula-flesch})-(\ref{formula-gulpease}) significa alterar os coeficientes $C_k$ de modo que os índices de legibilidade para os documentos nessa língua sejam os mais próximos possíveis daqueles obtidos na tradução para o inglês (ou italiano, no caso do índice Gulpease) do documento a partir das fórmulas com os coeficientes originais. Isso foi possível de ser realizado a partir de uma regressão linear múltipla usando o seguinte procedimento:

\begin{enumerate}
	\item Selecionamos uma amostra com $N = 100$ textos de diversos gêneros na Língua Portuguesa \cite{ListaTextos}.
	
	\item Usando os algoritmos da Seção \ref{algoritmos} escritos em JavaScript, calculamos todas as variáveis de interesse dos $N$ textos: $\left\lbrace x_1, x_2, \ldots, x_N\right\rbrace $ e $\left\lbrace y_1, y_2, \ldots, y_N\right\rbrace $.
	
	\item Obtemos as versões em inglês (ou italiano) dos $N$ textos.
	
	\item Obtemos os índices de legibilidade das versões traduzidas dos textos ($GL_k$, $k$-ésimo \textit{grade level}) usando ferramentais disponíveis na web, como o \textit{Readability Test Tool} \cite{RTT} e o \textit{Farfalla Project} \cite{Darian}.
	
	\item A partir dos dados anteriores, montamos uma matriz $N\times 3$, sendo a primeira coluna composta pelos valores de $x_k$, a segunda pelos valores de $y_k$, ambos obtidos no passo 2, e a terceira coluna, os índices das versões traduzidas dos textos, obtidos no passo 4.
	
	\item Obtemos os coeficientes $C_1$, $C_2$ e $C_3$ através da regressão linear múltipla dos dados da matriz formada no passo 5.
\end{enumerate}

Nas Seções seguintes apresentamos os resultados obtidos.

\subsection{Índice de Legibilidade de Flesch}
\label{indices-portugues}

Os resultados obtidos estão resumidos na tabela 1 e na figura 2.

\begin{table}[ht]
	\centering
	\caption{Coeficientes para o Índice de Legibilidade de Flesch obtidos via regressão linear múltipla.}
	\begin{tabular}{|lllll|}
		\hline
		\multicolumn{5}{|c|}{$R^2 = 0,890742$}                                                                                                               \\ \hline
		\multicolumn{1}{|l|}{}      & \multicolumn{1}{l|}{Referência}         & \multicolumn{1}{l|}{Valor}      & \multicolumn{1}{l|}{Erro padrão} & p-valor \\ \hline
		\multicolumn{1}{|l|}{$C_1$} & \multicolumn{1}{l|}{\textit{Intercept}} & \multicolumn{1}{l|}{226,614882} & \multicolumn{1}{l|}{8,744455}    & 0,00000 \\ \hline
		\multicolumn{1}{|l|}{$C_2$} & \multicolumn{1}{l|}{palavras/sentenças} & \multicolumn{1}{l|}{$-1{,}036134$}  & \multicolumn{1}{l|}{0,0930814}   & 0,00000 \\ \hline
		\multicolumn{1}{|l|}{$C_3$} & \multicolumn{1}{l|}{sílabas/palavras}   & \multicolumn{1}{l|}{$-72{,}451284$} & \multicolumn{1}{l|}{4,336399}    & 0,00000 \\ \hline
	\end{tabular}
\end{table}

\begin{figure}[ht]
	\centering
	\includegraphics[width=0.95\linewidth]{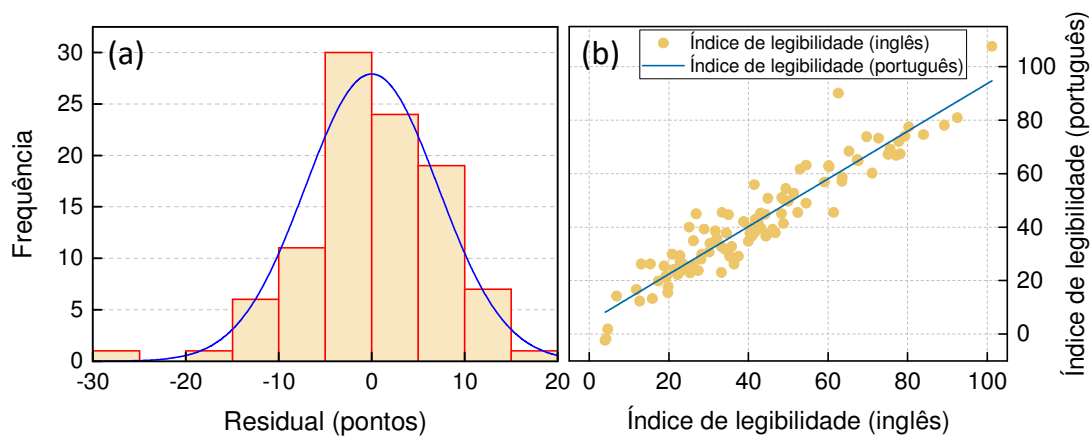}
	\caption{Resultado da regressão linear múltipla para o Índice de Legibilidade de Flesch. (a) Histograma da diferença entre o índice de legibilidade obtido (para textos em português, usando o software ALT) e o valor previsto (mesmo texto em inglês a partir da ferramenta RTT). A curva azul representa o ajuste da distribuição normal em torno dos dados do histograma. (b) Gráfico da dependência entre os índices de legibilidade obtido e previsto.}
	\label{fig2}
\end{figure}

Fórmula considerada para a Língua Portuguesa:

\begin{equation}
	\label{flesch-pt}
	\text{Índice de Legibilidade de Flesch} = 227 - 1{,}04 \times \left( \dfrac{\text{palavras}}{\text{sentenças}}\right) - 72 \times \left( \dfrac{\text{sílabas}}{\text{palavras}}\right) .
\end{equation}

Comentários:  Se a fórmula (\ref{flesch-pt}) for reaplicada nos 100 textos da amostra, 84 deles [ver Fig. \ref{fig2}(a)] apresentarão uma diferença no índice de legibilidade menor ou igual a 10 pontos em relação ao índice obtido com a fórmula original aplicada na versão em inglês do texto. A curva de ajuste azul obtida dos dados da Fig. \ref{fig2}(b) mostra que as variáveis $x$ e $y$ explicam 89,1\% da variância do índice de legibilidade.

\subsection{ Índice de Nebulosidade de Gunning }

Nossa adaptação para o Índice de Nebulosidade de Gunning faz uso de uma definição diferente de ``palavra complexa''. Enquanto que o Gunning fog index considera as palavras complexas como aquelas que contêm três ou mais sílabas, como apontado na Seção \ref{gunning-fog-index}, aqui fazemos uma checagem do vocábulo com um banco de palavras da Linguateca, um centro de recursos para processamento computacional da Língua Portuguesa \cite{Linguateca}. Essa modificação foi realizada com o intuito obter um método diferente no cálculo da legibilidade de um texto, já que outro índice, como o de Flesch-Kincaid, também usa o critério de contagem de sílabas para definir o que é ou não uma palavra complexa.

O banco de palavras usado está disponível em uma das URLs da Linguateca \cite{banco-palavras}, na opção ``Todos os corpos brasileiros'', coluna ``Lista de frequência total das formas no corpo''. Consideramos somente os cinco mil primeiros itens para formar o banco do programa ALT. O critério usado foi, então, considerar como ``palavra complexa'' todas aquelas além desse número.

Os resultados obtidos estão resumidos na tabela 2 e na figura 3.

\begin{table}[ht]
	\centering
	\caption{Coeficientes para o Índice de Nebulosidade de Gunning obtidos via regressão linear múltipla.}
	\begin{tabular}{|lllll|}
		\hline
		\multicolumn{5}{|c|}{$R^2 = 0,77333$}                                                                                                                       \\ \hline
		\multicolumn{1}{|l|}{}      & \multicolumn{1}{l|}{Referência}                  & \multicolumn{1}{l|}{Valor}    & \multicolumn{1}{l|}{Erro padrão} & p-valor \\ \hline
		\multicolumn{1}{|l|}{$C_1$} & \multicolumn{1}{l|}{\textit{Intercept}}          & \multicolumn{1}{l|}{1,00156}  & \multicolumn{1}{l|}{1,28036}     & 0,43599 \\ \hline
		\multicolumn{1}{|l|}{$C_2$} & \multicolumn{1}{l|}{palavras/sentenças}          & \multicolumn{1}{l|}{0,49261}  & \multicolumn{1}{l|}{0,02764}     & 0,00000 \\ \hline
		\multicolumn{1}{|l|}{$C_3$} & \multicolumn{1}{l|}{palavras complexas/palavras} & \multicolumn{1}{l|}{18,66057} & \multicolumn{1}{l|}{5,6943}      & 0,00146 \\ \hline
	\end{tabular}
\end{table}

\begin{figure}[ht]
	\centering
	\includegraphics[width=0.95\linewidth]{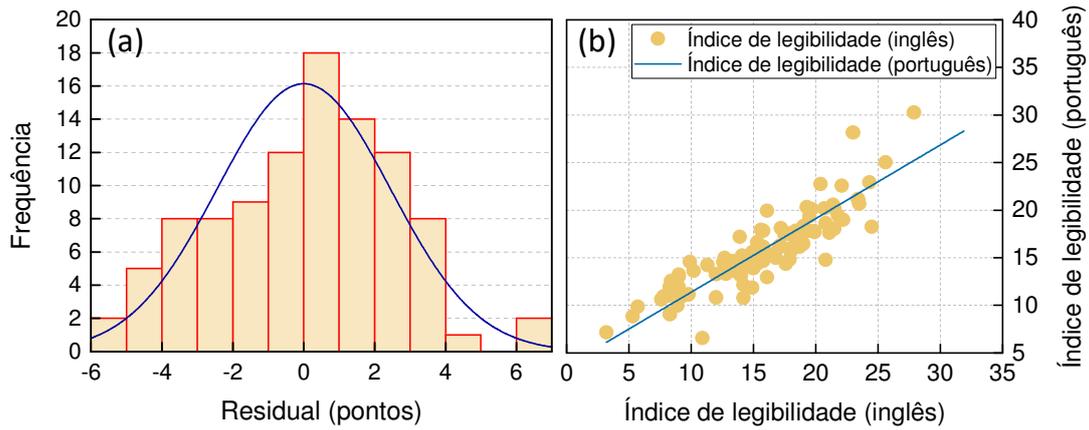}
	\caption{Resultado da regressão linear múltipla para o Índice de Nebulosidade de Gunning. (a)-(b): Idem Fig. \ref{fig2}.}
	\label{fig3}
\end{figure}

Fórmula considerada para a Língua Portuguesa:

\begin{equation}
	\label{gunning-fog-pt}
	\text{Índice de Nebulosidade de Gunning} = 0{,}49 \times \left( \dfrac{\text{palavras}}{\text{sentenças}}\right) + 19 \times \left( \dfrac{\text{palavras complexas}}{\text{palavras}}\right)
\end{equation}

Comentários: 

\begin{enumerate}
	\item O erro padrão no coeficiente $C_1$ (relativo ao \textit{intercept}) é maior do que o seu próprio valor. Ou seja, podemos considerar $C_1 = 0$ para todos os efeitos. Isso é corroborado pelo alto p-valor, bem maior do que o padrão normalmente adotado de 5\%. De fato, a fórmula original não possui esse coeficiente, o que indica uma boa adaptação para a língua portuguesa nesse ponto.
	
	\item Se a equação (\ref{gunning-fog-pt}) for reaplicada nos 100 textos da amostra, 73 deles [ver Fig. \ref{fig3}(a)] apresentarão uma diferença no índice de legibilidade menor ou igual a 3 pontos em relação ao índice obtido com a fórmula original aplicada na versão em inglês do texto. Este não é um resultado que podemos considerar interessante. Na média, isso quer dizer que 27\% dos textos analisados terão uma diferença no índice de mais de 3 pontos. Muito provavelmente isso vem da grande incerteza no coeficiente $C_3$, cujo erro padrão é quase um terço do seu valor. Esse coeficiente está ligado à variável palavras complexas/palavras. Isso pode ser um indicativo de que uma atualização no banco de palavras pode ser necessário.
	
	\item A curva de ajuste azul obtida dos dados da Fig. \ref{fig3}(b) mostra que as variáveis $x$ e $y$ explicam 77,3\% da variância do índice de Gunning. A grande incerteza no coeficiente $C_3$ explica esse valor relativamente baixo para $R^2$.
\end{enumerate}

\subsection{ Índice de Legibilidade Automatizado }

Os resultados obtidos estão resumidos na tabela 3 e na figura 4.

\begin{table}[ht]
	\centering
	\caption{Coeficientes para o ARI obtidos via regressão linear múltipla.}
	\begin{tabular}{|lllll|}
		\hline
		\multicolumn{5}{|c|}{$R^2 = 0,93696$}                                                                                                                          \\ \hline
		\multicolumn{1}{|l|}{}      & \multicolumn{1}{l|}{Referência}                  & \multicolumn{1}{l|}{Valor}       & \multicolumn{1}{l|}{Erro padrão} & p-valor \\ \hline
		\multicolumn{1}{|l|}{$C_1$} & \multicolumn{1}{l|}{\textit{Intercept}}          & \multicolumn{1}{l|}{$-$20,26065} & \multicolumn{1}{l|}{1,67994}     & 0,00000 \\ \hline
		\multicolumn{1}{|l|}{$C_2$} & \multicolumn{1}{l|}{letras/palavras}          & \multicolumn{1}{l|}{4,57058}     & \multicolumn{1}{l|}{0,36508}     & 0,00000 \\ \hline
		\multicolumn{1}{|l|}{$C_3$} & \multicolumn{1}{l|}{palavras/sentenças} & \multicolumn{1}{l|}{0,43664}     & \multicolumn{1}{l|}{0,01834}     & 0,00000 \\ \hline
	\end{tabular}
\end{table}

\begin{figure}[ht]
	\centering
	\includegraphics[width=0.95\linewidth]{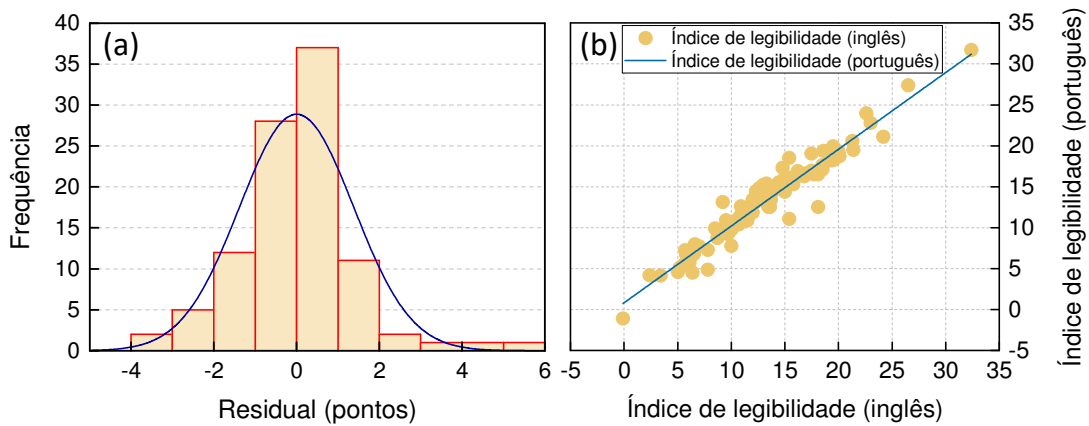}
	\caption{Resultado da regressão linear múltipla para o ARI.  (a)-(b): Idem Fig. \ref{fig2}.}
	\label{fig4}
\end{figure}

Fórmula considerada para a Língua Portuguesa:

\begin{equation}
	\label{ari-pt}
	\text{Índice de Legibilidade Automatizado} = 0{,}44 \times \left( \dfrac{\text{palavras}}{\text{sentenças}}\right)  + 4{,}6 \times \left( \dfrac{\text{caracteres}}{\text{palavras}}\right) - 20.
\end{equation}

Comentários: 

\begin{enumerate}
	\item Se a equação (\ref{ari-pt}) for reaplicada nos 100 textos da amostra, 88 deles [ver Fig. \ref{fig4}(a)] apresentarão uma diferença no índice de legibilidade menor ou igual a 2 pontos em relação ao índice obtido com a fórmula original aplicada na versão em inglês do texto, o que representa um excelente resultado. Esses fatos corroboram com os erros-padrão relativamente baixos nos três coeficientes e com o ótimo ajuste observado nos pontos da Fig. \ref{fig4}(b).
\end{enumerate}

\subsection{ Nível de Instrução de Flesch-Kincaid }

Os resultados obtidos estão resumidos na tabela 4 e na figura 5.

\begin{table}[ht]
	\centering
	\caption{Coeficientes para o Nível de Instrução de Flesch-Kincaid obtidos via regressão linear múltipla.}
	\begin{tabular}{|lllll|}
		\hline
		\multicolumn{5}{|c|}{$R^2 = 0,92273$}                                                                                                                 \\ \hline
		\multicolumn{1}{|l|}{}      & \multicolumn{1}{l|}{Referência}         & \multicolumn{1}{l|}{Valor}       & \multicolumn{1}{l|}{Erro padrão} & p-valor \\ \hline
		\multicolumn{1}{|l|}{$C_1$} & \multicolumn{1}{l|}{\textit{Intercept}} & \multicolumn{1}{l|}{$-18{,}11589$} & \multicolumn{1}{l|}{1,6077}      & 0,00000 \\ \hline
		\multicolumn{1}{|l|}{$C_2$} & \multicolumn{1}{l|}{palavras/sentenças} & \multicolumn{1}{l|}{0,36001}     & \multicolumn{1}{l|}{0,01712}     & 0,00000 \\ \hline
		\multicolumn{1}{|l|}{$C_3$} & \multicolumn{1}{l|}{sílabas/palavras}   & \multicolumn{1}{l|}{10,35177}    & \multicolumn{1}{l|}{0,79701}     & 0,00000 \\ \hline
	\end{tabular}
\end{table}

\begin{figure}[ht]
	\centering
	\includegraphics[width=0.95\linewidth]{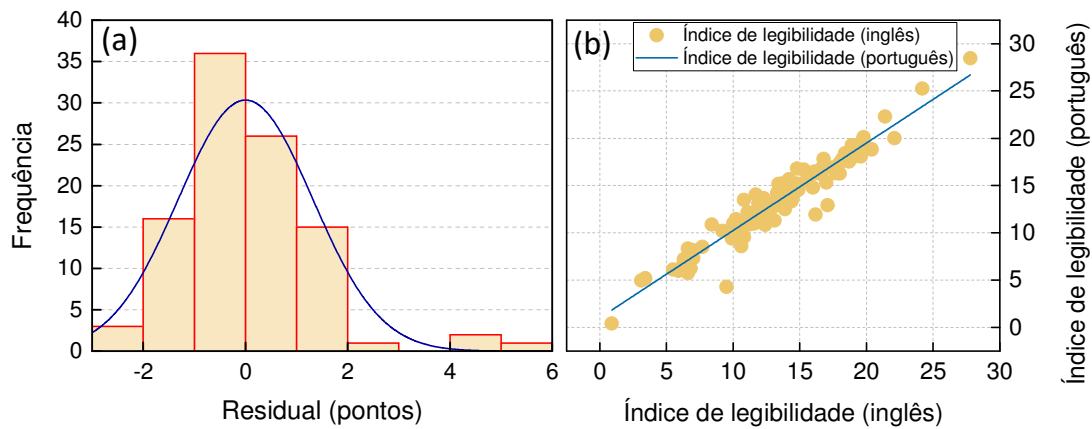}
	\caption{Resultado da regressão linear múltipla para o Nível de Instrução de Flesch-Kincaid.  (a)-(b): Idem Fig. \ref{fig2}.}
	\label{fig5}
\end{figure}

Fórmula considerada para a Língua Portuguesa:

\begin{equation}
	\label{flesch-kincaid-pt}
	\text{Nível de Instrução de Flesch-Kincaid} = 0{,}36 \times \left( \dfrac{\text{palavras}}{\text{sentenças}}\right) + 10{,}4 \times \left( \dfrac{\text{sílabas}}{\text{palavras}}\right) - 18.
\end{equation}

Comentários: 

\begin{enumerate}
	\item Tal como no ARI, temos aqui uma ótima adaptação indicada pelos baixos erros-padrão nos valores dos coeficientes em conjunto com um alto valor de $R^2$, conforme pode ser observado pelos resultados apresentados nas figuras \ref{fig5}(a) e \ref{fig5}(b).
\end{enumerate}

\newpage
\subsection{ Índice de Coleman-Liau }

Os resultados obtidos estão resumidos na tabela 5 e na figura 6.

\begin{table}[ht]
	\centering
	\caption{Coeficientes para o Índice de Coleman-Liau obtidos via regressão linear múltipla.}
	\begin{tabular}{|lllll|}
		\hline
		\multicolumn{5}{|c|}{$R^2 = 0{,}89221$}                                                                                                                \\ \hline
		\multicolumn{1}{|l|}{}      & \multicolumn{1}{l|}{Referência}         & \multicolumn{1}{l|}{Valor}     & \multicolumn{1}{l|}{Erro padrão} & p-valor \\ \hline
		\multicolumn{1}{|l|}{$C_1$} & \multicolumn{1}{l|}{\textit{Intercept}} & \multicolumn{1}{l|}{$-13{,}66302$} & \multicolumn{1}{l|}{1,61422}     & 0,00000 \\ \hline
		\multicolumn{1}{|l|}{$C_2$} & \multicolumn{1}{l|}{letras/palavras}    & \multicolumn{1}{l|}{5,39801}   & \multicolumn{1}{l|}{0,27242}     & 0,00000 \\ \hline
		\multicolumn{1}{|l|}{$C_3$} & \multicolumn{1}{l|}{sílabas/palavras}   & \multicolumn{1}{l|}{$-20{,}57984$}  & \multicolumn{1}{l|}{6,67523}     & 0,00000 \\ \hline
	\end{tabular}
\end{table}

\begin{figure}[ht]
	\centering
	\includegraphics[width=0.95\linewidth]{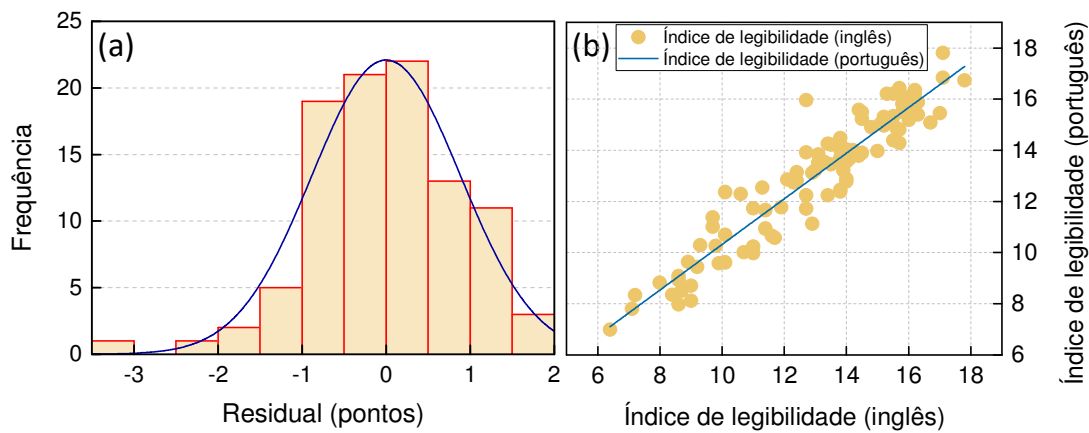}
	\caption{Resultado da regressão linear múltipla para o Índice de Coleman-Liau.  (a)-(b): Idem Fig. \ref{fig2}.}
	\label{fig6}
\end{figure}

Fórmula considerada para a Língua Portuguesa:

\begin{equation}
	\label{coleman-liau-pt}
	\text{Índice de Coleman-Liau} = 5{,}4 \times \left( \dfrac{\text{caracteres}}{\text{palavras}}\right) - 21 \times \left( \dfrac{\text{sentenças}}{\text{palavras}}\right) - 14.
\end{equation}

Comentários: 

\begin{enumerate}
	\item Tal como no ARI, temos aqui uma ótima adaptação indicada pelos baixos erros-padrão nos valores dos coeficientes em conjunto com um alto valor de $R^2$, conforme pode ser observado através dos resultados apresentados nas figuras \ref{fig6}(a) e \ref{fig6}(b).
\end{enumerate}

\subsection{ Índice Gulpease }

O Índice Gulpease não apresentou, dentro do limites das margens de erro, variações nos seus coeficientes quando aplicamos o procedimento destacado na Seção \ref{formulas-portugues}. Isso é um indicativo de que esse índice, na sua fórmula original para o italiano, pode ser usado para textos na Língua Portuguesa. Um motivo para isso pode ser a mesma origem latina desses dois idiomas.

\section{O software ALT}
\label{alt}

ALT -- Análise de Legibilidade Textual -- é um programa de computador capaz de retornar, em termos quantitativos, o nível de facilidade de leitura de textos. Ele foi desenvolvido pelos professores Marco Polo Moreno de Souza e Gleice Carvalho de Lima Moreno com a colaboração dos professores Nelson Hein e Adriana Kroenke Hein e escrito na linguagem JavaScript. Através dos algoritmos descritos na Seção \ref{algoritmos} e dos índices de legibilidade adaptados para a Língua Portuguesa, Eqs. (\ref{flesch-pt}) a (\ref{coleman-liau-pt}), ALT fornece o resultado final da legibilidade de um texto através da média aritmética de quatro índices que operam na escala de nível de instrução:

\begin{equation}
	\label{resultado}
	\text{Resultado} = \dfrac{1}{4} \left( \text{FK} + \text{GF} + \text{ARI} + \text{CL} \right),
\end{equation}

\noindent onde:

\begin{itemize}
	\item FK $=$ Nível de Instrução de Flesch-Kincaid,
	
	\item GF $=$ Índice de Nebulosidade de Gunning,
	
	\item ARI $=$ Índice de Legibilidade Automatizado e
	
	\item CL $=$ Índice de Coleman-Liau.
\end{itemize}

\noindent Não usamos para o resultado final os índices que operam na escala centígrada pelo fato de que uma conversão de escalas seria necessário. Entretanto, o programa ALT fornece as métricas individuais também para o Índice de Legibilidade de Flesch e para o Índice Gulpease, já que podem ser de interesse para o usuário.

\subsection{Visão geral}

\begin{figure}[ht]
	\centering
	\includegraphics[width=0.99\linewidth]{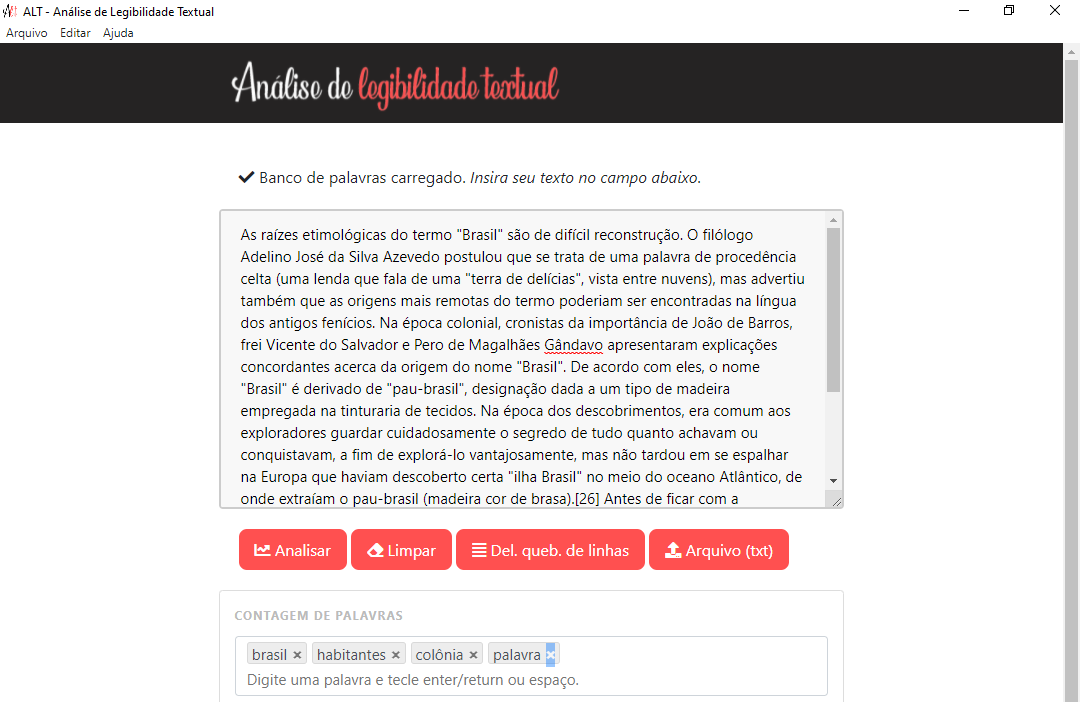}
	\caption{Parte do \textit{layout} do programa ALT na versão 1.1.0 para Windows.}
	\label{fig7}
\end{figure}

Apresentamos na Fig. \ref{fig7} o layout do programa ALT com a inserção da seção Etimologia do artigo Brasil do Wikipédia \cite{brasil}. Os índices de legibilidade são obtidos através do clique no botão \texttt{Analisar}, cujos resultados mostramos na próxima subseção.

\subsection{As métricas}

\begin{figure}[ht]
	\centering
	\includegraphics[width=0.99\linewidth]{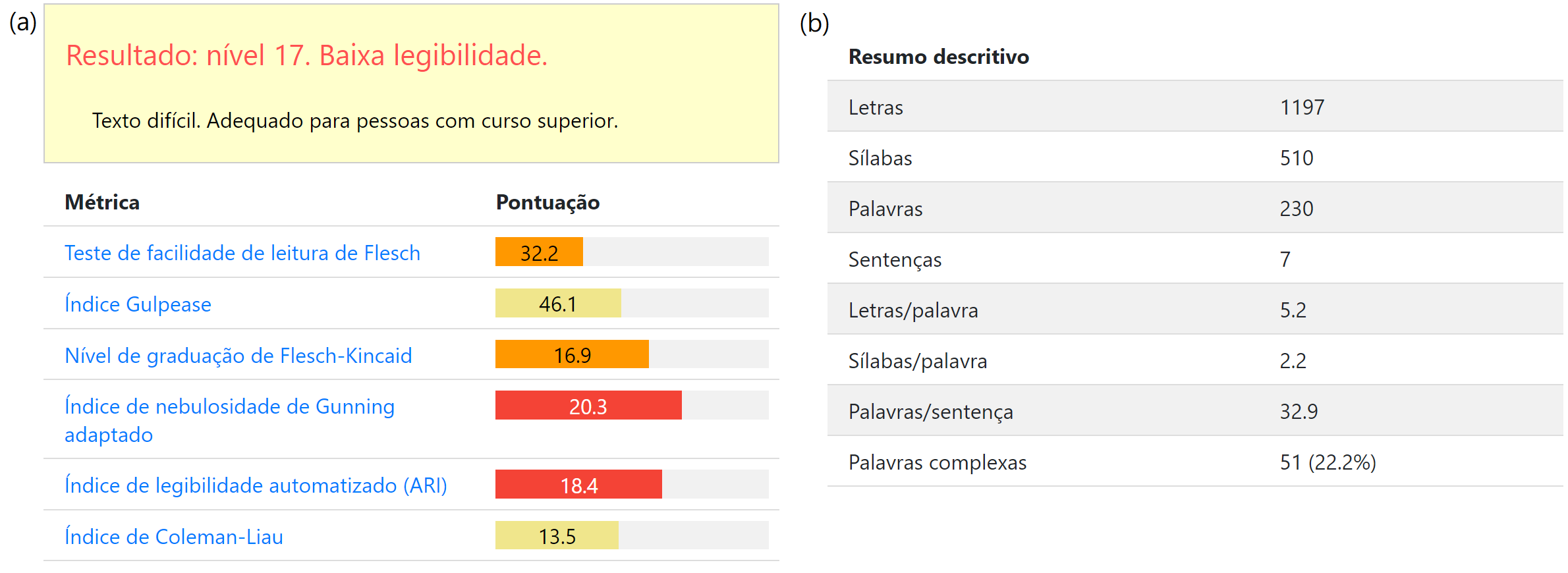}
	\caption{(a) Resultado final e índices de legibilidades específicos. (b) Resumo descritivo contendo as variáveis do texto analisado.}
	\label{fig8}
\end{figure}

A legibilidade do texto é informada em um campo amarelo, conforme pode ser visto na Fig. \ref{fig8}(a). O nível de legibilidade, obtido através da Eq. (\ref{resultado}), é um número que se situa, em geral, de 5 a 20. Além disso, apresentamos também a legibilidade em três graus: baixa, média e alta legibilidade, obtido através da seguinte receita:

\begin{itemize}
	\item Resultado abaixo de 13 pontos: alta legibilidade.
	
	\item Resultado a partir de 13 e abaixo de 17 pontos: média legibilidade.
	
	\item Resultado igual ou superior a 17 pontos: baixa legibilidade.
\end{itemize}

Logo abaixo do Resultado, o programa mostra os índices individuais captados através das seis métricas: Teste de facilidade de leitura de Flesch, Índice Gulpease, Nível de Instrução de Flesch-Kincaid, Índice de Nebulosidade de Gunning, Índice de Legibilidade Automatizado (ARI) e Índice de Coleman-Liau.

Por fim, o programa mostra todas as variáveis de interesse, como pode ser visto na Fig. \ref{fig8}(b): Número de letras, de sílabas, de palavras, de sentenças e de palavras complexas, e também algumas de suas razões: letras/palavra, sílabas/palavra e palavras/sentença.

\subsection{Busca por palavras específicas e a nuvem de palavras}

\begin{figure}[ht]
	\centering
	\includegraphics[width=0.99\linewidth]{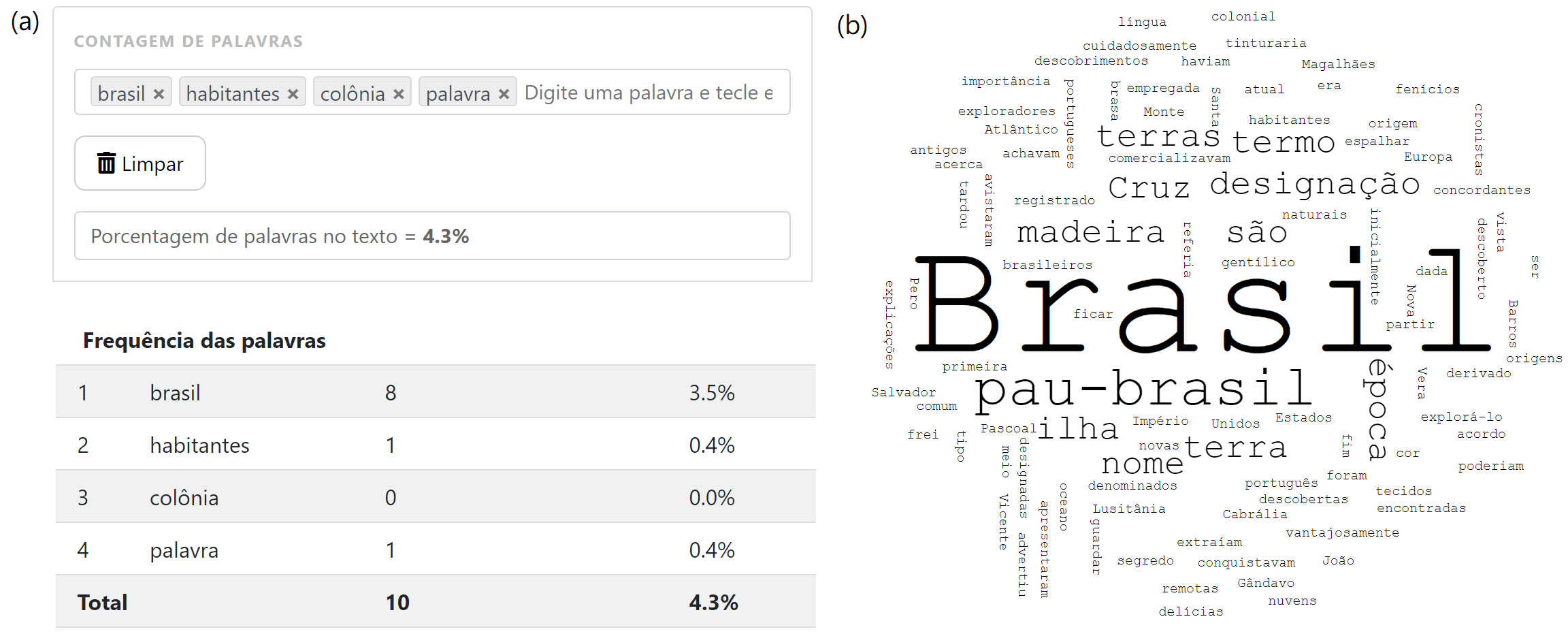}
	\caption{(a) Contagem de palavras específicas. (b) Nuvem de palavras.}
	\label{fig9}
\end{figure}

Outra parte do programa ALT é dedicada a analisar o conteúdo do texto em termos das palavras e de suas frequências. Logo abaixo do campo de inserção do texto, figuras \ref{fig7}, é possível procurar por palavras específicas dentro do texto. Uma vez que o botão \texttt{Analisar} é clicado, as frequências absolutas e relativas são apresentadas em uma tabela, conforme pode ser visto na figura \ref{fig9}(a).

Já a temática do texto pode ser inferida através de uma nuvem de palavras, onde uma imagem com palavras dispostas nos sentidos horizontal ou vertical são apresentadas com tamanhos proporcionais às suas frequências no texto. Essa é uma forma de análise visual de conteúdo, onde o teor do texto pode ser rapidamente obtido através das maiores palavras da nuvem. No texto-exemplo exposto, cuja nuvem de palavras é apresentada na Fig. \ref{fig9}(b), podemos observar claramente o destaque dos termos ``Brasil'', ``pau-brasil'', ``madeira'', ``Cruz'', ``termo'', dentre outros. Com essa informação, é possível saber do que se trata o texto, mesmo sem tê-lo lido: é uma discussão sobre a origem do nome do país Brasil.

É importante destacar que as palavras funcionais, que tem pouco papel na transmissão da informação semântica, foram removidas da nuvem, o que inclui as preposições, os artigos, os pronomes, as conjunções e as interjeições. Elas estão listadas no Apêndice \ref{lista-palavras-removidas}.

\subsection{Sugestões de melhorias}

Por fim, o programa ALT indica em um campo de texto alguns pontos que contribuem para um texto com baixa legibilidade. As sentenças consideradas longas (que consideramos como sendo aquelas formadas por 30 a 45 palavras) e muito longas (formadas por mais do que 45 palavras) ficam grifadas em amarelo e vermelho, respectivamente. Deixa-se como sugestão para que o usuário considere dividir as sentenças longas em duas e as sentenças muito longas em duas ou mais.

\begin{figure}[ht]
	\centering
	\includegraphics[width=0.75\linewidth]{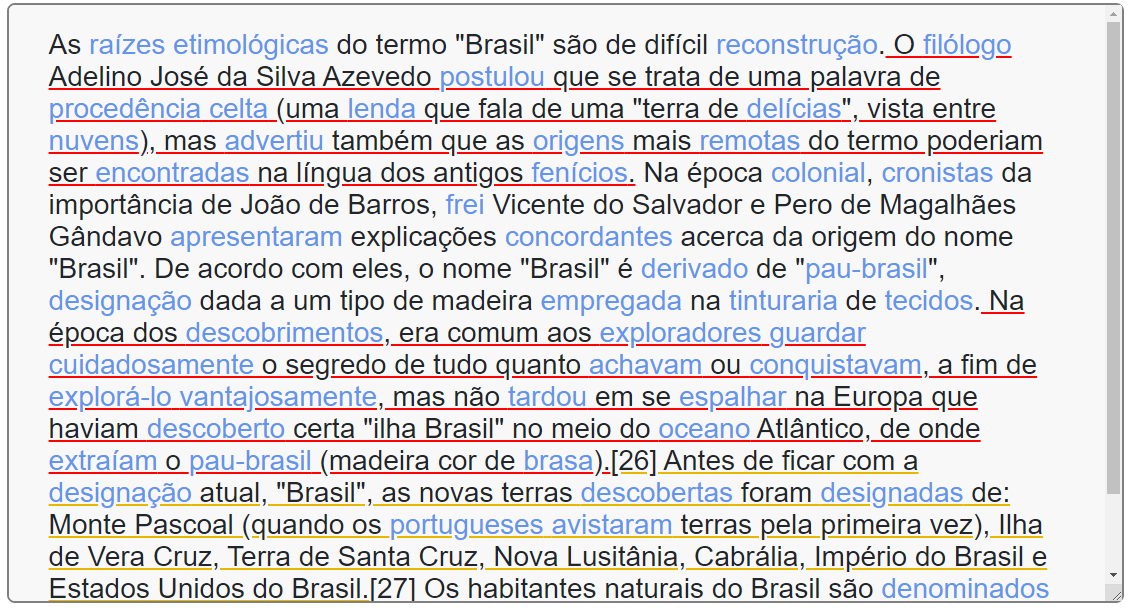}
	\caption{Apontamentos destacados pelo programa ALT. As sentenças consideradas longas ou muito longas estão grifadas em amarelo e vermelho, respectivamente. Já as palavras ``complexas'' estão indicadas pela fonte azul claro.}
	\label{fig10}
\end{figure}

Já as palavras consideradas ``complexas'' são destacadas pela cor azul claro. Deixa-se como sugestão substituí-las, a critério de cada usuário.

\subsection{Extensões}

Atualmente o software ALT está disponível em sete extensões, das quais uma deve ser acessada online enquanto que as outras seis devem ser baixadas e usadas no dispositivo do usuário. Essas últimas são

\begin{enumerate}
	\item \texttt{exe}: para instalação em dispositivos com Windows 64 bits.
	
	\item \texttt{apk}: para instalação em dispositivos com Android.
	
	\item \texttt{dmg}: para instalação em dispositivos com MacOS.
	
	\item \texttt{deb}: para instalação em dispositivos com Linux (algumas distribuições).
	
	\item \texttt{AppImage}: para uso (sem precisar de instalação) em dispositivos com Linux (algumas distribuições).
	
	\item \texttt{Play Store}: disponível a partir da loja de aplicativos Play Store para dispositivos Android.
\end{enumerate}

\begin{figure}[ht]
	\centering
	\includegraphics[width=0.45\linewidth]{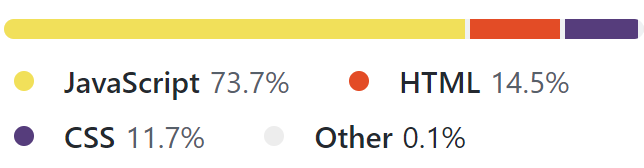}
	\caption{Composição das linguagens no código-fonte do programa ALT segundo o GitHub.}
	\label{fig11}
\end{figure}

O código-fonte do software ALT é composto pelas linguagens JavaScript, HTML e CSS, cujos percentuais de participação no projeto podem ser encontrados na Fig. \ref{fig11}. Todas as sete extensões são produzidas a partir de pequenas adaptações do mesmo projeto hospedado na plataforma GitHub. Para o desenvolvimento das extensões voltadas majoritariamente aos \textit{desktops} (exe, dmg, deb e AppImage), usamos o Electron js, um framework ligado ao Node.js e ao navegador Chromium. Já para o desenvolvimento das extensões direcionadas para os dispositivos móveis, usamos o Android Studio em conjunto sua componente WebView.

\section{Comparação das fórmulas adaptadas com suas versões originais na Língua Inglesa}
\label{comparacoes}

\begin{figure}[ht]
	\centering
	\includegraphics[width=0.99\linewidth]{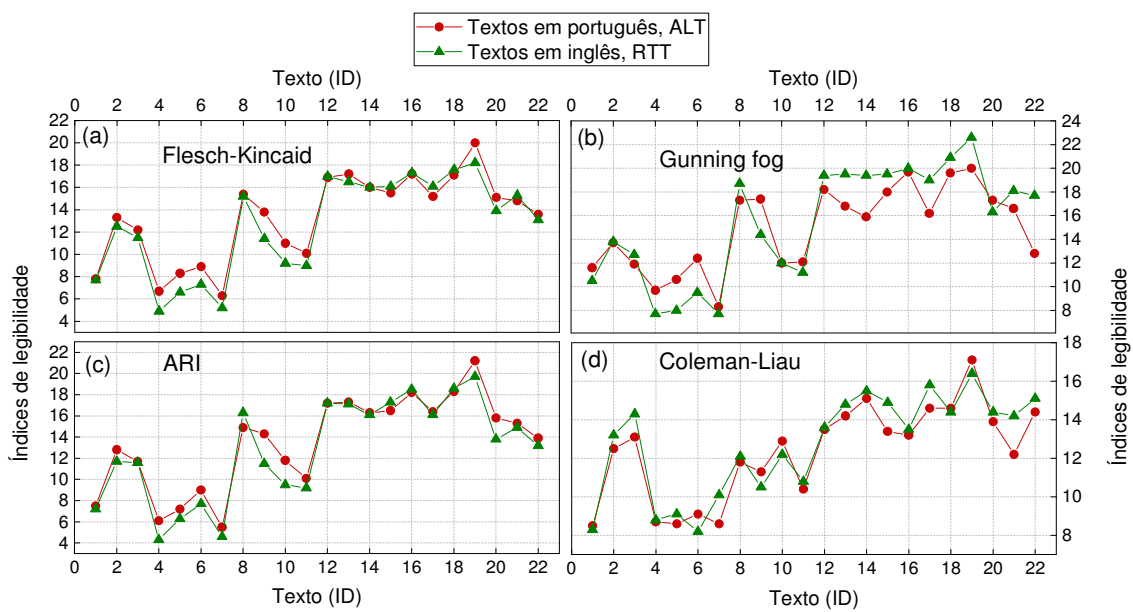}
	\caption{Comparação entre os índices de legibilidade dos textos do Apêndice \ref{tabela-textos-2}, representados pelo seu ID. Os círculos vermelhos representam os índices obtidos pelo programa ALT, ao passo que os triângulos verdes indicam os índices obtidos das versões em inglês dos textos, via ferramenta RTT.}
	\label{fig12}
\end{figure}

Nesta Seção, apresentamos uma análise comparativa entre os resultados obtidos pelo programa ALT, que faz uso das fórmulas de legibilidade adaptadas para a Língua Portuguesa [equações (\ref{flesch-pt}) a (\ref{coleman-liau-pt})], e aqueles obtidos da ferramenta \textit{Readability Test Tool} (RTT) \cite{RTT}, que usa as fórmulas originais para medir a legibilidade de textos na Língua Inglesa. A amostra consiste de 22 textos listados no Apêndice \ref{tabela-textos-2}. Os gráficos comparativos estão mostrados na Fig. \ref{fig12}, onde os círculos vermelhos indicam os índices de legibilidade obtidos pelo programa ALT enquanto que os triângulos verdes representam os índices obtidos da ferramenta RTT. Nitidamente, observamos uma forte correspondência entre os resultados obtidos, que é corroborada pelos dados da tabela \ref{correlacao}, onde temos a correlação de Pearson e a diferença média entre resultados obtidos dos dois programas. A ``margem de erro'' da diferença média foi considerada com dois desvios-padrões para cada lado, o que deve abranger em torno de 95\% dos textos de uma amostra qualquer.

Os resultados da Fig. \ref{fig12} e da Tab. \ref{correlacao} revelam uma menor qualidade na adaptação do Índice de Gunning fog, ainda que uma alta correlação ($> 0{,}9$) tenha sido obtida. Apesar da diferença média ter sido de apenas $-0{,}4$, a dispersão dos dados é mais do que o dobro de qualquer uma das diferenças médias obtidas para os outros índices. Isso indica que uma melhoria na adaptação do Índice de Gunning fog precise ser realizada para uma correspondência mais robusta.

\begin{table}[H]
	\centering
	\caption{Correlação e diferença média entre os resultados obtidos do programa ALT e da ferramenta RTT.}
	\label{correlacao}
	\begin{tabular}{|l|c|c|}
		\hline
		Índice          & Correlação & Diferença média   \\ \hline
		Flesch-Kincaid  & 98,0\%     & 0,7 $\pm$ 1,8     \\ \hline
		Gunning fog     & 91,3\%     & $-$ 0,4 $\pm$ 4,2 \\ \hline
		ARI             & 97,9\%     & 0,7 $\pm$ 2,0     \\ \hline
		Coleman-Liau    & 95,3\%     & $-0,4$ $\pm$ 1,6  \\ \hline
		Resultado final & 97,2\%     & 0,6 $\pm$ 2,0     \\ \hline
	\end{tabular}
\end{table}

\newpage
\section{Limitações das fórmulas de legibilidade textual}
\label{limitacoes}

Os índices de legibilidade precisam ser usados com bastante critério, já que nem sempre um índice baixo (na escala 0-20) indica um texto de fácil leitura. Observe, por exemplo, as primeiras proposições (da 1 até a 2.013) contidas no livro \textit{Tractatus Logico-Philosophicus}, de Ludwig Wittgenstein (Apêndice \ref{tractatus}).

Como é possível notar, trata-se de um texto filosófico que consiste em um conjunto de sentenças curtas agrupadas dentro de proposições numeradas, o que lhe garante índices de legibilidade abaixo dos 8 pontos nas quatro métricas do critério Nível de Instrução (\textit{Grade Level}): 

\begin{itemize}
	\item Flesch-Kincaid: 5,5
	
	\item Gunning fog: 6,6
	
	\item ARI: 4,3
	
	\item Coleman-Liau: 7,2,
\end{itemize}

\noindent com Resultado Final de 6 pontos. Isso indicaria, à princípio, que ele seria um texto adequado para crianças entre 11 e 12 anos. Entretanto, esta obra é considerada um texto de leitura bastante difícil, mesmo para especialistas \cite{marques}. Essa aparente incongruência surge porque as variáveis que caracterizam o texto são próximas daquelas encontradas em textos infantis: as variáveis letras/palavra, sílabas/palavra e palavras/sentença são 4,3, 1,9 e 10,3, respectivamente, e há apenas 8\% de palavras complexas. A complexidade do \textit{Tractatus} reside, dentre outros pontos, na necessidade de profunda reflexão e de muita familiaridade com conceitos de lógica moderna por parte do leitor. Esse é um bom exemplo de um texto que possui alta legibilidade ao mesmo tempo em que pode ser bem compreendido apenas por poucas pessoas fora do ambiente acadêmico especializado.

\begin{figure}[H]
	\centering
	\includegraphics[width=0.99\linewidth]{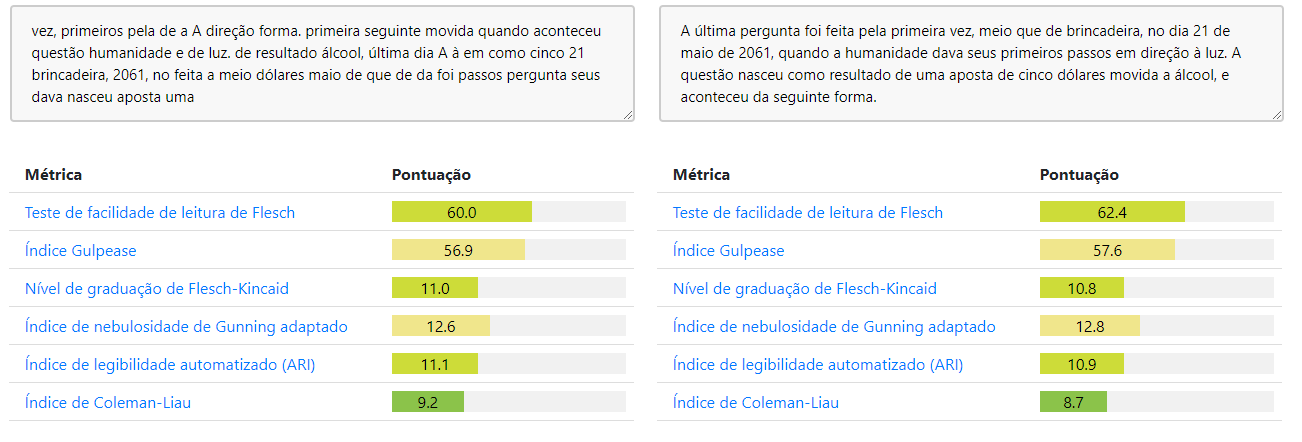}
	\caption{Versão embaralhada do primeiro parágrafo do conto ``A Última Pergunta'', de Issac Asimov (à esquerda) e sua versão original na tradução para o Português (à direita), com seus respectivos índices de legibilidade. O embaralhamento foi realizado através da ferramenta \textit{Word shuffler} \cite{word-shuffler}.}
	\label{fig13}
\end{figure}

Outro ponto que merece ênfase é que, naturalmente, os índices de legibilidade textual não analisam aspectos semânticos, sintáticos e pragmáticos. Observe, por exemplo, dois parágrafos e seus índices de legibilidade textual obtidos pelo programa ALT e apresentados na Fig. \ref{fig13}. Os índices devolvidos pelo programa são muito semelhantes, apesar de o parágrafo da esquerda não fazer sentido algum, já que ele é uma versão ``embaralhada'' do texto da direita. O parágrafo da esquerda é um caso de um texto que, mesmo incompreensível, possui alta legibilidade.

\section{Conclusões}
\label{conclusoes}
Este trabalho, influenciado pela teoria da ação comunicativa de Habermas, foi desenvolvido não para solucionar por completo as deficiências presentes na comunicação, mas para contribuir na redução de falhas que impedem o entendimento e a consequente credibilidade das comunicações escritas.

Então, para medir o grau de dificuldade na compreensão de um texto, foi construído o software ALT – Análise de Legibilidade Textual, ferramenta desenvolvida para a análise de legibilidade em Língua Portuguesa, disponível de forma livre na internet. Esta ferramenta foi construída para suprir duas necessidades: possibilitar a análise de legibilidade textual e preencher uma lacuna existente no ambiente científico.

Os índices de legibilidade usados foram os criados pelos pesquisadores para analisar a dificuldade no entendimento de um texto. Com a prova realizada por meio da correlação de Pearson, foi possível perceber uma boa adaptação dos índices de legibilidade. Logo, o software pode ser usado em pesquisas científicas e outros fins.

Os índices de legibilidade precisam ser usados com cuidado, uma vez que suas fórmulas usam apenas duas variáveis: palavras complexas (a mensagem atende a um grupo específico) e sentenças longas (a mensagem é difícil de ser entendida). Logo, não são capazes de medir a coesão e a coerência de uma comunicação escrita, que abrangem fatores semânticos, sintáticos e pragmáticos.





\appendix

\section{Lista de palavras removidas da nuvem}
\label{lista-palavras-removidas}

\begin{table}[H]
	\centering
	\caption{Lista de palavras removidas da nuvem}
	\begin{tabular}{|l|l|}
		\hline
		Preposições                                                                                       & \begin{tabular}[c]{@{}l@{}}a ante após até com contra de desde em entre\\  para perante por sem sob sobre trás\end{tabular}                                                                                                                                                                                                                                                                                                                                                                                                                                                                                                                                                                                                                                                                                                                                                                                                                                                                                                                                                                                                                                                                                                                                                                                                                                                                                                                                        \\ \hline
		\begin{tabular}[c]{@{}l@{}}Artigos, \\ artigos + pronomes e \\ artigos + preposições\end{tabular} & \begin{tabular}[c]{@{}l@{}}o a os as um uma uns umas àquele, àquela pelo pela pelos \\ pelas do da dos das no na nos nas de por para perante sem sob sobre \\ trás num nuns numa numas dum duns duma dumas à às\end{tabular}                                                                                                                                                                                                                                                                                                                                                                                                                                                                                                                                                                                                                                                                                                                                                                                                                                                                                                                                                                                                                                                                                                                                                                                                                                       \\ \hline
		Pronomes                                                                                          & \begin{tabular}[c]{@{}l@{}}eu, tu, ele, ela, nós, vós, eles, elas, meu, minha, meus, minhas, \\ teu, tua, teus, tuas, seu, sua, seus, suas, nosso, nossa, nossos,\\  nossas, vosso, vossa, vossos, vossas, este, esta, estes, estas, \\ esses, essas, esse, essa, aquele, aquela, aqueles, aquelas, aquilo, \\ tudo, toda, todas, todos, todas, algo, alguém, algum, alguma, \\ alguns, algumas, nada, ninguém, nenhum, nenhuma, nenhuns, \\ nenhumas, certo, certa, certos, certas, qualquer, quaisquer, \\ bastante, bastantes, mesmo, mesma, mesmo, mesmas, outra, \\ outro, outras, outros, qual, quais, cujo, cujas, cuja, cujos, que, \\ quanto, quantos, quantas, quanta, quem, onde, como, quando, \\ você, senhor, senhora, senhorita, Vossa Senhoria, Vossa Alteza, \\ Vossa Majestade, Vossa Mercê, Vossa Onipotência, Vossa \\ Excelência, vossa Magnificência, Vossa Santidade, Vossa \\ Reverendíssima, Vossa Eminência, o qual, a qual, os quais, \\ as quais, cujo, cuja, cujos, cujas, quanto, quanta, quantos, \\ quantas, me, mim, comigo, nós, conosco, te, ti, contigo, vos, \\ convosco, o, a, lhe, se, si, consigo, os, as, lhes vários, várias, \\ algo, cada, pouco, pouca, poucos e poucas, dele, deles, dela, \\ delas, deste, desta, destes, destas, desse, dessa, desses, dessas, \\ daquele, daquela, daqueles, daquelas, disto, disso, daquilo, \\ nesse, nessa, neste, nesta, nesses, nessas, nestes, nestas, naquilo\end{tabular} \\ \hline
		\begin{tabular}[c]{@{}l@{}}Conjunções \\ subordinadas \\ e coordenadas\end{tabular}               & \begin{tabular}[c]{@{}l@{}}porque, uma vez que, sendo que, visto que, como, tanto que, \\ sem que, de modo que, de forma que, de sorte que, tal qual, \\ do que , assim como, mais que, a menos que, conforme, \\ segundo, consoante, assim como, mesmo que, por mais que, \\ ainda que, se bem que, embora, se, caso, contanto que, \\ a menos que, sem que, salvo se, à medida que, \\ à proporção que, quanto mais, quanto maior, quanto menos, \\ quanto menor, quando, enquanto, sempre que, logo que, \\ depois que, pois que, visto que, uma vez que, porquanto, \\ já que, desde que, ainda que, posto que, por mais que, \\ apesar de que, antes que, a fim de que, para que, ao passo que, \\ quanto melhor, e, nem, mas também, como também, \\ bem como, mas, porém, todavia, contudo, entretanto, \\ no entanto, ou, ora, ora já, quer, já, logo, seja, talvez, \\ portanto, por isso, assim, por conseguinte, que, porque, \\ porquanto, pois, não só, não obstante\end{tabular}                                                                                                                                                                                                                                                                                                                                                                                                                                                                     \\ \hline
	\end{tabular}
\end{table}

\begin{table}[H]
	\centering
	\begin{tabular}{|l|l|}
		\hline
		\begin{tabular}[c]{@{}l@{}}Advérbios\\ (mais usados)\end{tabular} & \begin{tabular}[c]{@{}l@{}}certamente, deveras, efetivamente, decerto, \\ incontestavelmente, realmente, seguramente, sim porque, \\ porquê, conseguintemente, consequentemente, eis, acaso, \\ certamente, certo, decerto, porventura, possivelmente,\\  provavelmente, quiçá, talvez, apenas, exclusivamente, \\ salvo, senão, simplesmente, só, somente, unicamente, \\ ainda, até, inclusivamente, mesmo, nomeadamente, \\ também, apenas, assaz, bastante, bem, como, \\ completamente, demais, demasiado, demasiadamente, \\ excessivamente, extremamente, grandemente, \\ intensamente, levemente, ligeiramente, mais, mal, meio, \\ menos, mui, muito, nada, pouco, profundamente, quanto, \\ quão, quase, que, tanto, tão, todo, tudo, onde, como, \\ quando, porque, abaixo, acima, acolá, adiante, afora, aí, \\ além, algures, alhures, ali, antes, aonde, aquém, aqui, \\ atrás, através, avante, cá, debaixo, defronte, detrás, \\ dentro, diante, donde, fora, lá, longe, nenhures, onde, \\ perto, adrede, alerta, aliás, assim, bem, calmamente, \\ como, corajosamente, debalde, depressa, devagar, \\ dificilmente, felizmente, livremente, mal, melhor, pior, \\ principalmente, propositadamente, quase, selvaticamente, \\ sobremaneira, sobremodo, também, jamais, não, nem, \\ nunca, tampouco, antes, depois, primeiramente, \\ ultimamente, Quantidade:, algo, apenas, assaz, bastante, \\ demasiado, mais, menos, muito, nada, pouco, quanto, \\ quão, quase, tanto, tão, todo, tudo, afinal, agora, ainda, \\ amanhã, amiúde, anteontem, antes, antigamente, aqui, \\ breve, brevemente, cedo, comumente, \\ concomitantemente, dantes, depois, diariamente, \\ doravante, enfim, então, entrementes, finalmente, \\ hoje, imediatamente, já, jamais, logo, nunca, ontem, \\ ora, outrora, presentemente, primeiro, quando, \\ raramente, sempre, simultaneamente, tarde\end{tabular} \\ \hline
	\end{tabular}
\end{table}

\section{Tabela de textos da aplicação}
\label{tabela-textos-2}

\begin{table}[H]
	\centering
	\label{lista-textos-2}
	\caption{Lista de textos para comparação. As colunas FK (Flesch-Kincaid), GF (Gunning fog), ARI, CL (Coleman-Liau) e RF (Resultado Final) apresentam os índices de legibilidade dos textos representados pelo ID (identificador) obtidos pelo programa ALT e pela ferramenta \textit{Readability Test Tool} (números entre parênteses). }
	\begin{tabular}{|c|l|c|c|c|c|c|}
		\hline
		ID & Título                                                                                                & FK                                                    & GF                                                    & ARI                                                   & CL                                                    & RF                                                \\ \hline
		1  & \begin{tabular}[c]{@{}l@{}}Dom Casmurro, caps. 1 e 2\\ (M. de Assis)\end{tabular}                     & \begin{tabular}[c]{@{}c@{}}7.8\\ (7.7)\end{tabular}   & \begin{tabular}[c]{@{}c@{}}11.6\\ (10.5)\end{tabular} & \begin{tabular}[c]{@{}c@{}}7.5\\ (7.2)\end{tabular}   & \begin{tabular}[c]{@{}c@{}}8.5\\ (8.3)\end{tabular}   & \begin{tabular}[c]{@{}c@{}}9\\ (8)\end{tabular}   \\ \hline
		2  & Notícia Portal G1                                                                                     & \begin{tabular}[c]{@{}c@{}}13.3\\ (12.5)\end{tabular} & \begin{tabular}[c]{@{}c@{}}13.7\\ (13.8)\end{tabular} & \begin{tabular}[c]{@{}c@{}}12.8\\ (11.7)\end{tabular} & \begin{tabular}[c]{@{}c@{}}12.5\\ (13.2)\end{tabular} & \begin{tabular}[c]{@{}c@{}}13\\ (13)\end{tabular} \\ \hline
		3  & Jornal GGN                                                                                            & \begin{tabular}[c]{@{}c@{}}12.2\\ (11.5)\end{tabular} & \begin{tabular}[c]{@{}c@{}}11.9\\ (12.7)\end{tabular} & \begin{tabular}[c]{@{}c@{}}11.7\\ (11.6)\end{tabular} & \begin{tabular}[c]{@{}c@{}}13.1\\ (14.3)\end{tabular} & \begin{tabular}[c]{@{}c@{}}12\\ (12)\end{tabular} \\ \hline
		4  & O Patinho feio                                                                                        & \begin{tabular}[c]{@{}c@{}}6.7\\ (4.9)\end{tabular}   & \begin{tabular}[c]{@{}c@{}}9.7\\ (7.7)\end{tabular}   & \begin{tabular}[c]{@{}c@{}}6.1\\ (4.3)\end{tabular}   & \begin{tabular}[c]{@{}c@{}}8.7\\ (8.8)\end{tabular}   & \begin{tabular}[c]{@{}c@{}}8\\ (6)\end{tabular}   \\ \hline
		5  & Pinóquio                                                                                              & \begin{tabular}[c]{@{}c@{}}8.3\\ (6.6)\end{tabular}   & \begin{tabular}[c]{@{}c@{}}10.6\\ (8)\end{tabular}    & \begin{tabular}[c]{@{}c@{}}7.2\\ (6.3)\end{tabular}   & \begin{tabular}[c]{@{}c@{}}8.6\\ (9.1)\end{tabular}   & \begin{tabular}[c]{@{}c@{}}9\\ (7)\end{tabular}   \\ \hline
		6  & João e Maria                                                                                          & \begin{tabular}[c]{@{}c@{}}8.9\\ (7.3)\end{tabular}   & \begin{tabular}[c]{@{}c@{}}12.4\\ (9.5)\end{tabular}  & \begin{tabular}[c]{@{}c@{}}9.0\\ (7.7)\end{tabular}   & \begin{tabular}[c]{@{}c@{}}9.1\\ (8.2)\end{tabular}   & \begin{tabular}[c]{@{}c@{}}10\\ (8)\end{tabular}  \\ \hline
		7  & \begin{tabular}[c]{@{}l@{}}Vida de Droga, págs. 5, 6 e 7\\ (Walcyr Carrasco)\end{tabular}             & \begin{tabular}[c]{@{}c@{}}6.3\\ (5.2)\end{tabular}   & \begin{tabular}[c]{@{}c@{}}8.3\\ (7.7)\end{tabular}   & \begin{tabular}[c]{@{}c@{}}5.5\\ (4.6)\end{tabular}   & \begin{tabular}[c]{@{}c@{}}8.6\\ (10.1)\end{tabular}  & \begin{tabular}[c]{@{}c@{}}7\\ (7)\end{tabular}   \\ \hline
		8  & \begin{tabular}[c]{@{}l@{}}Contabilidade Rural, págs. 61-63\\ (J. C. Marion)\end{tabular}             & \begin{tabular}[c]{@{}c@{}}15.4\\ (15.2)\end{tabular} & \begin{tabular}[c]{@{}c@{}}17.3\\ (18.7)\end{tabular} & \begin{tabular}[c]{@{}c@{}}14.9\\ (16.3)\end{tabular} & \begin{tabular}[c]{@{}c@{}}11.8\\ (12.1)\end{tabular} & \begin{tabular}[c]{@{}c@{}}15\\ (15)\end{tabular} \\ \hline
		9  & \begin{tabular}[c]{@{}l@{}}Fund. de Física Vol. 1, pág. 5\\ (Halliday)\end{tabular}                   & \begin{tabular}[c]{@{}c@{}}13.8\\ (11.4)\end{tabular} & \begin{tabular}[c]{@{}c@{}}17.4\\ (14.4)\end{tabular} & \begin{tabular}[c]{@{}c@{}}14.3\\ (11.5)\end{tabular} & \begin{tabular}[c]{@{}c@{}}11.3\\ (10.5)\end{tabular} & \begin{tabular}[c]{@{}c@{}}14\\ (12)\end{tabular} \\ \hline
		10 & \begin{tabular}[c]{@{}l@{}}C\# e .Net, págs. 3 e 4\\ (J. E. Saraiva)\end{tabular}                     & \begin{tabular}[c]{@{}c@{}}11.0\\ (9.2)\end{tabular}  & \begin{tabular}[c]{@{}c@{}}12.0\\ (12.0)\end{tabular} & \begin{tabular}[c]{@{}c@{}}11.8\\ (9.5)\end{tabular}  & \begin{tabular}[c]{@{}c@{}}12.9\\ (12.2)\end{tabular} & \begin{tabular}[c]{@{}c@{}}12\\ (10)\end{tabular} \\ \hline
		11 & \begin{tabular}[c]{@{}l@{}}O Senhor dos Anéis, Vol. 1, \\ cap. 1, 3 prim. pág. (Tolkien)\end{tabular} & \begin{tabular}[c]{@{}c@{}}10.1\\ (9.0)\end{tabular}  & \begin{tabular}[c]{@{}c@{}}12.1\\ (11.2)\end{tabular} & \begin{tabular}[c]{@{}c@{}}10.1\\ (9.2)\end{tabular}  & \begin{tabular}[c]{@{}c@{}}10.4\\ (10.8)\end{tabular} & \begin{tabular}[c]{@{}c@{}}11\\ (10)\end{tabular} \\ \hline
		12 & \begin{tabular}[c]{@{}l@{}}Artigo Rev. Educação e \\ Pesquisa (Introdução)\end{tabular}               & \begin{tabular}[c]{@{}c@{}}16.9\\ (17.0)\end{tabular} & \begin{tabular}[c]{@{}c@{}}18.2\\ (19.4)\end{tabular} & \begin{tabular}[c]{@{}c@{}}17.2\\ (17.2)\end{tabular} & \begin{tabular}[c]{@{}c@{}}13.5\\ (13.6)\end{tabular} & \begin{tabular}[c]{@{}c@{}}16\\ (16)\end{tabular} \\ \hline
		13 & \begin{tabular}[c]{@{}l@{}}Artigo Rev. Saúde e Debate \\ (Introdução)\end{tabular}                    & \begin{tabular}[c]{@{}c@{}}17.2\\ (16.5)\end{tabular} & \begin{tabular}[c]{@{}c@{}}16.8\\ (19.5)\end{tabular} & \begin{tabular}[c]{@{}c@{}}17.3\\ (17.1)\end{tabular} & \begin{tabular}[c]{@{}c@{}}14.2\\ (14.8)\end{tabular} & \begin{tabular}[c]{@{}c@{}}16\\ (17)\end{tabular} \\ \hline
		14 & \begin{tabular}[c]{@{}l@{}}Artigo Rev. Ensino de Física \\ (Introdução)\end{tabular}                  & \begin{tabular}[c]{@{}c@{}}16.0\\ (16.0)\end{tabular} & \begin{tabular}[c]{@{}c@{}}15.9\\ (19.4)\end{tabular} & \begin{tabular}[c]{@{}c@{}}16.3\\ (16.1)\end{tabular} & \begin{tabular}[c]{@{}c@{}}15.1\\ (15.5)\end{tabular} & \begin{tabular}[c]{@{}c@{}}16\\ (16)\end{tabular} \\ \hline
		15 & \begin{tabular}[c]{@{}l@{}}Artigo Rev. Physis \\ (Arquivos)\end{tabular}                              & \begin{tabular}[c]{@{}c@{}}15.5\\ (16.1)\end{tabular} & \begin{tabular}[c]{@{}c@{}}18.0\\ (19.5)\end{tabular} & \begin{tabular}[c]{@{}c@{}}16.5\\ (17.3)\end{tabular} & \begin{tabular}[c]{@{}c@{}}13.4\\ (14.9)\end{tabular} & \begin{tabular}[c]{@{}c@{}}16\\ (16)\end{tabular} \\ \hline
		16 & \begin{tabular}[c]{@{}l@{}}Artigo Rev. Est. Teo.\\ Psicanalítica (Introdução)\end{tabular}            & \begin{tabular}[c]{@{}c@{}}17.2\\ (17.3)\end{tabular} & \begin{tabular}[c]{@{}c@{}}19.7\\ (20)\end{tabular}   & \begin{tabular}[c]{@{}c@{}}18.2\\ (18.5)\end{tabular} & \begin{tabular}[c]{@{}c@{}}13.2\\ (13.5)\end{tabular} & \begin{tabular}[c]{@{}c@{}}17\\ (17)\end{tabular} \\ \hline
		17 & \begin{tabular}[c]{@{}l@{}}Artigo Rev. Cont. \\ Contemp. (Introdução)\end{tabular}                    & \begin{tabular}[c]{@{}c@{}}15.2\\ (16.1)\end{tabular} & \begin{tabular}[c]{@{}c@{}}16.2\\ (19.0)\end{tabular} & \begin{tabular}[c]{@{}c@{}}16.4\\ (16.1)\end{tabular} & \begin{tabular}[c]{@{}c@{}}14.6\\ (15.8)\end{tabular} & \begin{tabular}[c]{@{}c@{}}16\\ (17)\end{tabular} \\ \hline
		18 & \begin{tabular}[c]{@{}l@{}}Artigo Rev. Direito e Praxis \\ (Introdução)\end{tabular}                  & \begin{tabular}[c]{@{}c@{}}17.1\\ (17.6)\end{tabular} & \begin{tabular}[c]{@{}c@{}}19.6\\ (20.9)\end{tabular} & \begin{tabular}[c]{@{}c@{}}18.3\\ (18.6)\end{tabular} & \begin{tabular}[c]{@{}c@{}}14.6\\ (14.4)\end{tabular} & \begin{tabular}[c]{@{}c@{}}17\\ (17)\end{tabular} \\ \hline
		19 & \begin{tabular}[c]{@{}l@{}}Fís. Atômica e Conhec. Humano, \\ págs. 85-87 (N. Bohr)\end{tabular}       & \begin{tabular}[c]{@{}c@{}}20.0\\ (18.2)\end{tabular} & \begin{tabular}[c]{@{}c@{}}20.0\\ (22.6)\end{tabular} & \begin{tabular}[c]{@{}c@{}}21.2\\ (19.7)\end{tabular} & \begin{tabular}[c]{@{}c@{}}17.1\\ (16.4)\end{tabular} & \begin{tabular}[c]{@{}c@{}}20\\ (19)\end{tabular} \\ \hline
		20 & \begin{tabular}[c]{@{}l@{}}Relat. Sustentabilidade \\ Coca-Cola, pág. 41\end{tabular}                 & \begin{tabular}[c]{@{}c@{}}15.1\\ (13.9)\end{tabular} & \begin{tabular}[c]{@{}c@{}}17.3\\ (16.3)\end{tabular} & \begin{tabular}[c]{@{}c@{}}15.8\\ (13.8)\end{tabular} & \begin{tabular}[c]{@{}c@{}}13.9\\ (14.4)\end{tabular} & \begin{tabular}[c]{@{}c@{}}16\\ (14)\end{tabular} \\ \hline
		21 & Relat. Gerdau, pág. 17                                                                                & \begin{tabular}[c]{@{}c@{}}14.8\\ (15.3)\end{tabular} & \begin{tabular}[c]{@{}c@{}}16.6\\ (18.1)\end{tabular} & \begin{tabular}[c]{@{}c@{}}15.3\\ (14.9)\end{tabular} & \begin{tabular}[c]{@{}c@{}}12.2\\ (14.2)\end{tabular} & \begin{tabular}[c]{@{}c@{}}15\\ (15)\end{tabular} \\ \hline
		22 & Relat. Itaú 2019, pág. 28                                                                             & \begin{tabular}[c]{@{}c@{}}13.6\\ (13.1)\end{tabular} & \begin{tabular}[c]{@{}c@{}}12.8\\ (17.7)\end{tabular} & \begin{tabular}[c]{@{}c@{}}13.9\\ (13.2)\end{tabular} & \begin{tabular}[c]{@{}c@{}}14.4\\ (15.1)\end{tabular} & \begin{tabular}[c]{@{}c@{}}14\\ (14)\end{tabular} \\ \hline
	\end{tabular}
\end{table}

\newpage
\section{\textit{Tractatus Logico-Philosophicus}, primeiras proposições}
\label{tractatus}

\begin{quotation}
	1* O mundo é tudo o que ocorre.
	
	1.1 O mundo é a totalidade dos fatos, não das coisas.
	
	1.11 O mundo é determinado pelos fatos e por isto consistir em todos os fatos.
	
	1.12 A totalidade dos fatos determina, pois, o que ocorre e também tudo que não ocorre.
	
	1.13 Os fatos, no espaço lógico, são o mundo.
	
	1.2 O mundo se resolve em fatos.
	
	1.21 Algo pode ocorrer ou não ocorrer e todo o resto permanecer na mesma.
	
	2 O que ocorre, o fato, é o subsistir dos estados de coisas.
	
	2.01 O estado de coisas é uma ligação de objetos (coisas).
	
	2.011 É essencial para a coisa poder ser parte constituinte der estado de coisas.
	
	2.012 Nada é acidental na lógica: se uma coisa puder aparecer num estado de coisas, a possibilidade do estado de coisas já deve estar antecipada nela.
	
	2.0121 Parece, por assim di.zer, acidental que à coisa, que poderia subsistir sozinha e para si, viesse ajustar-se em seguida uma situação.
	
Se as coisas podem aparecer em estados de coisas, então isto já. deve estar nelas.
	
	(Algo lógico não pode ser meramente-possível. A lógica trata de cada possibilidade e tôdas as possibilidades são fatos quê lhe pertencem.)
	
Assim como não podemos pensar objetos espaciais fora do espaço, os temporais fora do tempo, assim não podemos pensar nenhum objeto fora da possibilidade de sua ligação com outros.
	
Se posso pensar o objeto ligando-o ao estado de coisas, não posso então pensá-lo fora da possibilidade dessa ligação.

2.0123 Se conheço o objeto, também conheço tôdas as possibilidades de seu aparecer em estados de coisas.

(Cada uma dessas possibilidades deve estar na natureza do objeto.)

Não é possível posteriormente encontrar nova possibilidade.
	
	2.0122 coisa é autônoma enquanto puder aparecer em tôdas as situações possíveis, mas esta forma de
	autonomia é uma forma de conexão com o estado de coisas, uma forma de heteronomia. (É impossível
	palavras comparecerem de dois modos diferentes, sôzinhas e na proposição.)
	
	2.01231 Para conhecer um objeto não devo com efeito conhecer suas propriedades externas — mas tôdas
	as internas.
	
	2.0124 Ao serem dados todos os objetos, dão-se também todos os possíveis estados de coisas.
	
	2.013 coisa está como num espaço de estados de coisas possíveis. Posso pensar êste espaço vazio,
	mas não a. coisa sem o espaço.
\end{quotation}

\end{document}